\pdfoutput=1
\documentclass[preprint,10pt]{elsarticle} 
\usepackage{graphicx,subcaption}
\usepackage{natbib}
\usepackage{colortbl}
\usepackage{amssymb}
\usepackage{moreverb} % JAHO
 \usepackage{float}   % JAHO
\usepackage{listings}
\usepackage[linesnumbered,ruled]{algorithm2e}
\SetCommentSty{mycommfont}
\usepackage{nomencl}
\usepackage[english]{babel}
\usepackage{nomencl}  % it doesn't compile !!!! See MISCELL.doc 
\usepackage{leftidx} % For attaching left-hand sided scripts
 \usepackage{amsmath, amsthm, amssymb}

\usepackage[left=1.5cm,top=2cm,right=1.5cm,bottom=2.5cm]{geometry} % 
\usepackage{color}
\usepackage{listings}
\usepackage{mathrsfs}

\usepackage{multirow}
\usepackage{color,colordvi,graphicx,xcolor}
\journal{Axioms}

\definecolor{mygreen}{rgb}{0,0.6,0}
\definecolor{mygray}{rgb}{0.5,0.5,0.5}
\definecolor{mymauve}{rgb}{0.58,0,0.82}
\usepackage{hyperref}
\makenomenclature

\usepackage{lipsum}  
\usepackage{tikz}
\usepackage{bbm}

\usepackage{import}
\usepackage{xifthen}
\usepackage{pdfpages}
\usepackage{transparent}

\newcommand{\norm}[1]{\left\lVert#1\right\rVert} %added by Raul
\usepackage{optidef} %added by Raul
\usepackage{tcolorbox}

\newtcolorbox{mybox}[1][]{
    colback=white,       % Background color
    #1                   % Additional options
}

\newcommand{\PhiInf}{\boldsymbol{\Phi}}
\newcommand{\PhiSup}{\bar{\boldsymbol{\Phi}}}
\newcommand{\qInf}{\mathbf{q}}
\newcommand{\XiInf}{\boldsymbol{\Xi}}
\newcommand{\XiSup}{\bar{\boldsymbol{\Xi}}}
\begin{document}

\begin{frontmatter}

\title{A discrete physics-informed training for projection-based \\ reduced order models with neural networks}

\author[upc]{N. Sibuet}
\author[upc,cimne]{S. Ares de Parga\corref{cor1}}
\ead{sebastian.ares@upc.edu}
\author[upc,cimne]{J.R. Bravo}
\author[upc,cimne]{R. Rossi}
% \author[rvt]{}

%
\cortext[cor1]{Corresponding author}

\address[upc]{Universitat Polit\`{e}cnica de Catalunya, Department of Civil and Environmental Engineering (DECA), Barcelona, Spain}
\address[cimne]{Centre Internacional de M\`{e}todes Num\`{e}rics en Enginyeria (CIMNE), Barcelona, Spain}
% \address[upc-terrasa]{Universitat Polit\`{e}cnica de Catalunya,  E.S. d'Enginyeries Industrial, Aeroespacial i Audiovisual de Terrassa (ESEIAAT), Terrassa, Spain}%

\begin{abstract}
%\lipsum[2-4]

This paper presents a physics-informed training framework for projection-based Reduced Order Models (ROMs). We extend the PROM-ANN architecture by complementing snapshot-based training with a FEM-based, discrete physics-informed residual loss, bridging the gap between traditional projection-based ROMs and physics-informed neural networks (PINNs). Unlike conventional PINNs that rely on analytical PDEs, our approach leverages FEM residuals to guide the learning of the ROM approximation manifold. Key contributions include: (1) a parameter-agnostic, discrete residual loss applicable to non-linear problems, (2) an architectural modification to PROM-ANN improving accuracy for fast-decaying singular values, and (3) an empirical study on the proposed physics informed training process for ROMs.

The method is demonstrated on a non-linear hyperelasticity problem, simulating a rubber cantilever under multi-axial loads. The main accomplishment in regards to the proposed residual-based loss is its applicability on non-linear problems by interfacing with FEM software while maintaining reasonable training times. The modified PROM-ANN outperforms POD by orders of magnitude in snapshot reconstruction accuracy, while the original formulation is not able to learn a proper mapping for this use-case. Finally, the application of physics informed training in ANN-PROM modestly narrows the gap between data reconstruction and ROM accuracy, however it highlights the untapped potential of the proposed residual-driven optimization for future ROM development. This work underscores the critical role of FEM residuals in ROM construction and calls for further exploration on architectures beyond PROM-ANN.

\end{abstract}
\begin{keyword}
Reduced Order Model (ROM), Physics Informed Neural Networks (PINNs), Artificial Neural Network (ANN), Projection-Based Model Reduction, Proper Orthogonal Decomposition (POD)\\[1.0em]

\textbf{MSC:} 65M60, 68T07
\end{keyword}
\end{frontmatter}

\section{Introduction}

In recent years, the development of increasingly sophisticated high-fidelity models (HFMs) is crucial for simulating complex physical phenomena. One of the main computational challenges in HFMs is the need for very fine spatio-temporal resolutions, which results in extremely high-dimensional problems that can take weeks to solve, even on numerous parallelized computing cores. Therefore, it is vital to create methods that significantly reduce the computational time and resources required to enable the application of these models in time-sensitive scenarios, including real-time control systems, simulation-driven optimization and digital twins \cite{brunton2021data,jiang2021industrial,Ares2024parallel}.

To address these computational challenges, projection-based reduced-order models (ROMs) have emerged as a powerful strategy, allowing for rapid simulation by approximating the dynamics of the high-dimensional system within a lower-dimensional latent space \cite{hesthaven2016certified}. ROMs can be broadly classified as \textit{intrusive} or \textit{non-intrusive}, depending on how the reduced equations are derived. Intrusive ROMs project the governing equations directly onto a reduced space, typically via Galerkin projection \cite{quarteroni2014reduced}, Least-Squares Petrov–Galerkin (LSPG) projection \cite{Carlberg2011, Carlberg2013, Carlberg2017, grimberg2021mesh}, or related Petrov–Galerkin formulations \cite{Ares2023}. In contrast, non-intrusive ROMs avoid direct equation manipulation, instead employing supervised-learning methods like artificial neural networks (ANNs) or Gaussian processes to map parameters directly to reduced solutions, thereby fully decoupling the online computation from the original high-fidelity solver \cite{hesthaven2018non, wang2019non, guo2018reduced, guo2019data, casenave2015nonintrusive}.

Intrusive projection-based ROMs typically involve an offline-online decomposition. In the offline stage, a latent space is computed from a collection of high-fidelity solutions sampled over a defined parametric domain. Proper Orthogonal Decomposition (POD) is a common approach, relying on Singular Value Decomposition (SVD) to obtain an optimal linear basis that efficiently captures the dominant dynamics \cite{sirovich_turbulence_1987, Holmes_Lumley_Berkooz_1996, eckart1936approximation}. The online stage then involves solving a reduced optimization problem within this latent space, resulting in dramatically reduced computational costs compared to the original HFM. Such strategies have demonstrated effectiveness in parametric linear problems \cite{Antoulas2005, Cuong05} and nonlinear problems with moderate complexities \cite{Rowley2004, Benner2015}.

Non-intrusive ROMs have attracted attention for scenarios where intrusive access to HFM operators is impractical or unavailable. Early successes include Gaussian processes and ANN regression models for aeroacoustic and structural problems \cite{casenave2015nonintrusive, guo2018reduced}, as well as neural networks for predicting flow coefficients in combustion dynamics \cite{hesthaven2018non, wang2019non}. Recent extensions leverage convolutional autoencoders for complex spatio-temporal predictions \cite{xu2020multi} and generalized data-driven frameworks for nonlinear PDEs \cite{guo2019data}. Later work with graph neural networks \cite{pichi_graph_2023} aims at bypassing the need for structured meshes that convolutional autoencoders have. Despite their advantages in computational efficiency, these methods require large training datasets and their scale is still directly tied to the HFM's dimensions.

Nevertheless, intrusive linear ROMs, particularly those using POD-Galerkin formulations, have been noted to exhibit accuracy degradation and stability issues in highly nonlinear or convection-dominated regimes. For example, spurious mode growth was reported in compressible flows \cite{rowley2004model}, stability issues have been examined from dynamical systems perspectives \cite{iollo2000stability}, and convergence difficulties have been highlighted in reactive flow scenarios \cite{huang2018challenges}. These limitations have been attributed to the slow decay of the Kolmogorov n-width \cite{pinkus_n-widths_1985}, fundamentally restricting the representational power of linear subspaces.

Consequently, a growing body of research has focused on developing nonlinear ROM strategies to overcome these fundamental expressivity barriers. Within intrusive ROMs, the formal theory for generic non-linear projection-based ROM was introduced in \cite{lee2020model}, which was first demonstrated using a convolutional neural network to define a nonlinear mapping between the full and latent spaces. Despite several limitations—namely, limited scalability and the need for structured meshes—this architecture paved the way for further non-linear architectures. These include methods based on quadratic manifolds \cite{barnett2022quadratic}, piecewise linear manifolds via local POD approaches \cite{amsallem2012nonlinear,grimberg2021mesh,bravo2024subspace}, and the PROM-ANN architecture from \cite{BARNETT2023112420,chmiel2024assessment}; the latter essentially being a non-linear approximation of POD using dense neural networks. 

Another approach to deal with non-linearizable dynamics is the use of spectral submanifolds \cite{li_data-free_2025,cenedese_data-driven_2022}, which are more mathematically sound than data-based ones. These type of non-linear manifolds can be found under certain non-resonance conditions \cite{haller_nonlinear_2016}, and their direct computation requires explicit knowledge of nonlinear coefficients in the equations of motion. Recent developments in \cite{li_data-free_2025} present a non-intrusive and data-free method to obtain these spectral submanifolds, but are currently limited to mechanical systems with up to cubic order non-linearities.

In this paper, we propose a novel residual-informed training approach for constructing nonlinear ROM operators. While existing projection-based ROMs construct their projection operators based exclusively on learning the reconstruction of solution snapshots, we propose incorporating the discrete residual of the FEM-based HFM—the same residual used during online ROM projection—into the training of these operators. The reason being that, since the residual is the quantity to be optimized during ROM inference, its more accurate representation will necessarily improve the solutions of the ROMs.

Our methodology is inspired by Physics-Informed Neural Networks (PINNs) \cite{raissi_physics_2017} and their variants, e.g. \cite{cuomo2022scientific, li_physics-informed_2024}. PINNs are models based on neural networks that learn the solution function to a system of partial differential equations (PDEs) in their continuous form. This is achieved by embedding the governing PDEs directly into the loss function and taking advantage of the automatic differentiation capabilities of common machine learning libraries. However, their application in engineering workflows can be hindered by their difficulties to cope with: irregular or discontinuous domains, complex PDEs due to spectral bias (which hinders their ability to handle high-frequency terms \cite{wang_when_2022}), discontinuities (such as shock waves \cite{fuks_limitations_2020}), and plasticity (requiring significant effort to develop workarounds for these issues \cite{niu_modeling_2023}).

In contrast, the Finite Element Method (FEM) \cite{zienkiewicz_finite_1971}, along with other HFMs like Finite Differences or Finite Volumes Methods, remains the gold standard for solving complex physical behaviors, particularly in engineering applications. This has led to recent efforts to bridge FEM and PINNs. For example, \cite{Meethal2023} and \cite{Thang2023} proposed discrete PINNs that use FEM to compute residuals and their derivatives for backpropagation in linear problems. These approaches predict nodal values using neural networks that take simulation parameters as the only input. Meanwhile, the spatial discretization and boundary conditions are implicitly integrated via the FEM residual. While \cite{Thang2023} focuses exclusively on structural mechanics, \cite{Meethal2023} generalizes the approach but still only for linear PDEs. Another intermediate approach is presented in \cite{eshaghi_variational_2025}, which introduces a physics-informed neural operator inference \cite{wang_learning_2021} framework that takes a discretized, variational form of the PDE as the training loss. This variational form is based on the energy of the system and is closely related to the FEM one, although not generalizable to the same degree. They remark how traditional PINNs only enforce the strong form of the PDE on the collocation points, resulting in lack of global smoothness, while a variational approach ensures this implicitly.

Other semi-intrusive strategies have been proposed that embed reduced-order residual information into neural networks to enhance generalization and enforce physical consistency in parametric regimes. We refer to these as semi-intrusive methods, since they require partial access to governing equations or their projections during training—such as reduced residuals—but retain a decoupled, non-intrusive structure at inference time. In \cite{chen2021physics}, a Physics-Reinforced Neural Network (PRNN) was introduced, which minimizes a hybrid loss composed of reduced residuals and projection data in the latent space. Subsequently, \cite{hijazi2023pod} combined POD-Galerkin ROMs with a neural network trained on both data and the residual of the reduced Navier–Stokes equations, enabling the same architecture to address forward prediction and inverse parameter-identification tasks. Moreover, ~\cite{brivio2024ptpi} enhanced POD–DL-ROMs by incorporating a strong-form PDE residual into the training process and adopting a pre-train plus fine-tune strategy that significantly reduces computational cost in nonlinear flow problems.

Beyond reduced-order modeling, broader physics-informed machine learning frameworks such as the Deep Galerkin Method for high-dimensional free-boundary PDEs \cite{sirignano2018dgm} and PINN-based RANS modeling for turbulent incompressible flows \cite{eivazi2022physics} further demonstrate the growing applicability of physics-aware neural architectures to complex nonlinear systems. Related efforts have explored manifold-learning and structure-preserving neural architectures outside traditional ROM contexts, including shallow masked autoencoders for nonlinear mechanical simulations~\cite{kim2022fast}, mechanics-informed neural networks for robust constitutive modeling~\cite{as2022mechanics}, physics-informed architectures for modeling mistuned integrally bladed rotors~\cite{kelly2025physics}, FE-ROM-informed neural operators for structural dynamics~\cite{yang2025fe}, and two-tier deep neural architectures for multiscale nonlinear mechanics~\cite{hong2024physics}. Although these methods primarily remain data-driven, their built-in physics constraints substantially enhance predictive capability. Despite these promising developments, a gap remains in fully leveraging the discrete physics of the HFM directly during the training of projection-based ROMs, particularly within nonlinear manifold approximations.

In this work, we propose a method to incorporate physics information into the ROM non-linear approximation manifold by using the FEM residual as the training loss. This requires two key components: (1) a flexible ROM architecture capable of accommodating custom loss functions for the definition of the projection operators, and (2) a framework to integrate FEM residuals into the training process. For the architecture, we adapt the PROM-ANN framework from \cite{BARNETT2023112420,chmiel2024assessment}, which uses neural networks in a scalable manner. We modify this architecture to suit our needs and develop a framework to integrate FEM residuals into a neural network's loss, along with their Jacobians for the back-propagation. Our approach assumes the user has access to a fully functional FEM software capable of providing these quantities —a reasonable requirement given the need for an HFM to generate training data and perform intrusive ROM.

Our residual-based loss function differs from previous studies in its generality and flexibility. Instead of minimizing the FEM residual norm directly, we compute the difference between the obtained residual and a ground truth, thereby learning the behavior of the residual in non-converged conditions. We also provide an optimized implementation for computing the gradient of the residual loss, enabling efficient backpropagation.

To validate our methodology, we apply it to a steady-state structural mechanics problem involving a cantilever modeled by an unstructured mesh. The FEM software of choice is KratosMultiphysics \cite{vicente_mataix_ferrandiz_2024_14185721}, which we adapted for this purpose. Our results show that the modified PROM-ANN architecture, combined with the residual-based loss, achieves slightly but consistently improved accuracy in ROM simulations. We do acknowledge that training with the residual loss is computationally expensive compared to the traditional snapshot-based loss. To address this, we propose using residual training only as a fine-tuning step after initial snapshot-based training, significantly reducing overall training time.

While adapting the architecture of PROM-ANN to suit our specific needs, we also identified opportunities to enhance the original design, enabling it to learn effectively across a broader range of scenarios, even without incorporating the residual-based loss. These improvements are presented as additional contributions in this paper.

It should be noted that, although projection-based ROMs have also been employed to accelerate inverse problems~\cite{brigham2009inverse, frangos2010surrogate}, the present work exclusively addresses forward parametric simulations. Extensions to inverse tasks—such as parameter estimation—would require further developments, for instance involving adjoint-based gradients or parameter-to-output mappings, which lie beyond the scope of this study.

In summary, our contributions are threefold:

\begin{enumerate}
    \item \textbf{Physics-Informed Residual-Based Loss}: A residual-based loss function is introduced for training ROMs using discrete FEM residuals. While generally applicable to non-linear problems and ROM architectures, it is demonstrated here within the PROM-ANN framework. The loss is parameter-agnostic and integrated through a general backpropagation strategy compatible with existing FEM infrastructure.

    \item \textbf{Enhanced PROM-ANN Architecture}: Modifications are proposed to improve the original PROM-ANN framework, enabling it to handle problems with fast-decaying singular-value spectra. These include scaling strategies that enhance training stability and general applicability.

    \item \textbf{Quantitative Evaluation of Residual Training}: A comprehensive study is conducted to assess the impact of residual-informed training on both snapshot reconstruction and ROM simulation, demonstrating modest but consistent improvements and laying the groundwork for future refinements.
\end{enumerate}

The application of the physics-informed training in this specific architecture does not yield enough enhancement in terms of accuracy to justify the increase in training time. However, our findings suggest that focusing on residual behavior could unlock further potential in non-linear ROMs. Particularly when combined with architectures specifically designed for this purpose. Possible directions to make the training process faster and to make the effect of the physics-based training more impactful are proposed within this paper.

The rest of the paper is organized as follows. Section \ref{sec:background} provides a review of the main methods that form the basis for this work, i.e. PINNs and projection-based ROM with an emphasis on the original PROM-ANN architecture. In Section \ref{sec:discrete_pinn} we propose a discrete, FEM-based residual loss and develop an adequate implementation strategy for it. Then, Section \ref{sec:modification_to_PROM-ANN_architecture} provides a series of modifications to the original PROM-ANN architecture and loss that make it compatible with problems with fast-decaying singular values. Section \ref{sec:integration_of_physics-based_loss_into_PROM-ANN} effectively merges the developments presented in the two previous sections to enable physics-informed non-linear ROM. Further on, Section \ref{sec: usecase and evaluation methodology} explains the specific FEM problem in which the developments will be tested, as well as the software of choice and the neural-network training strategy. Section \ref{sec:results} shows the results of applying the methods developed throughout the paper onto our specific use-case. Section \ref{sec:discussion_and_future_work} further discusses the implications of the results from the previous section and proposes further research directions for future improvements. Finally, \ref{sec:conclusions} closes the paper with the most significant conclusions.
\section{Background}
\label{sec:background}

\subsection{Physics-Informed Neural Networks}
\label{subsec: PINNS}

In this section we introduce PINNs, highlighting the most relevant aspects to our case. Our focus is on PINNs for forward parametric problems. Comprehensive reviews on the numerous variations of PINNs can be found in  e.g. \cite{cuomo2022scientific,cai2021physics}

Consider a physical problem in the following form 

\begin{equation}
\label{eq:formal_pde_definition}
\left\{
\begin{aligned}
     \mathcal{R}(\boldsymbol{u}(\boldsymbol{z}; \boldsymbol{\mu}),\boldsymbol{\mu}) &= \boldsymbol{0}  &  \hspace{5mm} &\boldsymbol{z} \in \Omega \times (0,T]  \ , \\
     \mathcal{B} (\boldsymbol{u}(\boldsymbol{z}; \boldsymbol{\mu}) , \boldsymbol{z}) &= g(\boldsymbol{z} ,\boldsymbol{\mu})  & &\boldsymbol{z} \in \partial\Omega \times (0,T] \ ,  \\
     \mathcal{I}(\boldsymbol{u} (\boldsymbol{z};\boldsymbol{\mu}), \boldsymbol{z}) &= f (\boldsymbol{z},\boldsymbol{\mu}) &  \hspace{5mm} &\boldsymbol{z} \in \Omega \times \{0\} \ , 
\end{aligned}
\right.
\end{equation}

\noindent where $\Omega \subset \mathbb{R}^d$ represents the spatial domain with boundary $\partial \Omega$, $\boldsymbol{u}: \Omega \times \mathbb{R}_+ \times \mathbb{R}^p \rightarrow \mathbb{R}^{N_{\text{\tiny dof}}}$ is the unknown solution, $\boldsymbol{z}$ is a spatio-temporal coordinates vector, $\boldsymbol{\mu} \in \mathcal{P} \subset \mathbb{R}^p$ is the parameters vector encapsulating geometrical, material, and/or initial and boundary conditions. $\mathcal{R}(\cdot)$ is the differential operator in residual form encapsulating the physics. $\mathcal{B}(\cdot)$ is an operator indicating boundary conditions, and $\mathcal{I}(\cdot)$ is an operator indicating initial conditions. 

In PINNs one uses a neural network for representing the solution function, as

\begin{equation}
    \boldsymbol{u} \approx \mathcal{N}(\boldsymbol{z} , \boldsymbol{\mu} ; \boldsymbol{\Theta})  \ ,
    \label{eq: NN as func}
\end{equation}

\noindent where \(\mathcal{N}\) denotes the neural network, and \(\boldsymbol{\Theta}\) encapsulates its learnable parameters. The optimal parameters for the neural network are obtained by minimizing a loss function:

\begin{equation}
    \boldsymbol{\Theta}^* = \underset{\boldsymbol{\Theta}}{\text{arg min}} \ \mathcal{L}(\boldsymbol{\Theta})  \ ,
\end{equation}

\noindent where the loss function $\mathcal{L}(\boldsymbol{\Theta})$ can be splitted into
\begin{equation}
    \mathcal{L}(\boldsymbol{\Theta}) = \omega_{\mathcal{R}} \mathcal{L}_{\mathcal{R}}(\boldsymbol{\Theta}) + \omega_{d}\mathcal{L}_{d}(\boldsymbol{\Theta})  \ .  
    \label{eq: loss PINNs}
\end{equation}

Here, \(\omega_{d}\) and \(\omega_{\mathcal{R}}\) are weighting factors that balance the contribution of the data-driven loss \(\mathcal{L}_{d}\) and the residual loss \(\mathcal{L}_{\mathcal{R}}\), respectively, to the overall loss function. The data-driven loss \(\mathcal{L}_{d}\) is related to the mismatch between \(mN_d\) ground-truth data points \(\boldsymbol{u}_{ij}^*\), and the output of the PINN for the same spatio-temporal and parametric coordinates. This data-driven loss \(\mathcal{L}_{d}\) reads

\begin{equation}
    \mathcal{L}_{d} = \text{MSE}_d = \frac{1}{m N_d} \sum_{j=1}^{m} \sum_{i=1}^{N_d} \norm{\mathcal{N}(\boldsymbol{z}_i , \boldsymbol{\mu}_j ; \boldsymbol{\Theta}) - \boldsymbol{u}^*_{ij} }^{2} \ ,
    \label{eq: loss data pinn}
\end{equation}

\noindent where \( N_d\) stands for the number of collocation points considered in the spatio-temporal domain. Moreover, \(m\) represents the number of points considered in the parametric space \(\mathcal{P}\). 

The residual loss \(\mathcal{L}_{\mathcal{R}}\) relates to the substitution in the residual operator, as defined in Eq.~\eqref{eq:formal_pde_definition}, of the unknown solution with its neural network approximation as defined in Eq.~\eqref{eq: NN as func}, and reads

\begin{equation}
    \mathcal{L}_{\mathcal{R}} = \text{MSE}_\mathcal{R} =  \frac{1}{m N_\mathcal{R}} \sum_{j=1}^{m} \sum_{i=1}^{N_\mathcal{R}} \norm{ \mathcal{R}(\mathcal{N}(\boldsymbol{z}_i , \boldsymbol{\mu}_j ; \boldsymbol{\Theta}) ,\boldsymbol{\mu}_j) }^{2}  \ ,
    \label{eq: Loss residual pinn}
\end{equation}

\noindent where \( N_\mathcal{R} \) represents the number of residual evaluation points, which might be different from the collocation points used in the data-driven loss. Admittedly, from the substitution of Eq.~\eqref{eq: NN as func} into Eq.~\eqref{eq:formal_pde_definition}, two more terms could be considered, corresponding to the initial and boundary conditions. However, here we assume that the initial and boundary conditions have been embedded into the PINN, as in \cite{ZHU201956,SUN2020112732}, therefore ensuring their compliance, and not entering in the loss function.

The fact that the PINN itself is a differentiable mapping from spatio-temporal coordinates to the corresponding solution means that gradients of the solution and temporal derivatives of it are directly computable within the deep learning software via auto-differentiation. This allows for the internal computation of the loss and its gradient.

\subsection{Projection-Based Reduced Order Models} 

\subsubsection{The Full Order Model (FOM)}

In the initial formulation of our problem, we represented the physics using a continuous residual form, as shown in Eq.~\eqref{eq:formal_pde_definition}, revisited here for clarity:

\begin{equation}
\label{eq:residual_definition}
\mathcal{R}(\boldsymbol{u}(\boldsymbol{z}; \boldsymbol{\mu});\boldsymbol{\mu}) = \boldsymbol{0}  \hspace{5mm} \boldsymbol{z} \in \Omega \times (0,T] \ .
\end{equation}

To adapt this continuous form for computational analysis, we discretize the domain using FEM. By applying FEM discretization to Eq.~\eqref{eq:residual_definition} and incorporating initial and boundary conditions, we arrive at a set of governing equations in the form of a $\boldsymbol{\mu}$-parametric operator $\textbf{R}:\mathbb{R}^{N} \times \mathcal{P} \rightarrow \mathbb{R}^N$. This results in the following discrete representation:
\begin{equation}
    \textbf{R}(\textbf{u}; \boldsymbol{\mu}) = \boldsymbol{0} \ ,
    \label{eq: residual discrete}
\end{equation}

\noindent where $\textbf{u}  \in \mathbb{R}^{N}$ is the FOM solution vector containing the value of the solution on every degree of freedom of the spatial discretization, and $\boldsymbol{\mu} \in \mathcal{P} \subset \mathbb{R}^p$ is the parameters vector encapsulating e.g. geometrical variations, material properties or boundary conditions.

Given an initial guess of the FOM solution at a step $t$, symbolically represented as $\mathbf{u}_t^0(\boldsymbol{\mu})$\footnote{For our current discussion can be either a time step, or a loading step following a predefined trajectory}, the actual solution corresponding to the step can be obtained via an iterative method (e.g. Newton's) like
\begin{subequations}
\begin{align}
- \textbf{J}(\textbf{u}_{t}^k(\boldsymbol{\mu}) ; \boldsymbol{\mu}) \delta \textbf{u}_t^k(\boldsymbol{\mu}) & = \textbf{R} (\textbf{u}_{t}^{k}(\boldsymbol{\mu}) ; \boldsymbol{\mu})\\
\textbf{u}^{k+1}_t (\boldsymbol{\mu}) & = \textbf{u}_t^{k}(\boldsymbol{\mu}) + \delta\textbf{u}_t^k (\boldsymbol{\mu})
\end{align}
\end{subequations}

\noindent where $\textbf{J} = \frac{\partial \textbf{R}}{\partial \textbf{u}}$ is a jacobian matrix (or an approximation thereof), $k$ is the current non-linear iteration, and $\delta \mathbf{u}_t^k \in \mathbb{R}^{N}$ is the solution state increment. 

\noindent
In practice, the FEM discretization can yield millions of degrees of freedom, making each nonlinear iteration prohibitively expensive. This challenge grows acute in scenarios demanding rapid solutions (e.g., design optimization, digital twins). To address this, various surrogate models have emerged, ranging from intrusive methods that directly manipulate the governing equations to non-intrusive black-box approaches. This work builds upon an intrusive, projection-based reduced-order model.

\subsection{Manifold Projection-Based ROMs}
\label{subsec: manifold-projection-based-rom}

Following the standard procedure introduced in \cite{lee2020model} for non-linear projection-based ROMs (also known as manifold-based ROMs), we begin by approximating the FOM's solution state variable $\mathbf{u}^*(\boldsymbol{\mu})$ through a general decoder function defined as:
\begin{equation}
\mathbf{u}^*(\boldsymbol{\mu}) \approx \mathbf{u}(\boldsymbol{\mu}) = D_u(\mathbf{q}(\boldsymbol{\mu})),
\label{eq: Genral Decoder}
\end{equation}
where $D_u: \mathbb{R}^n \rightarrow \mathbb{R}^{N}$ maps a reduced solution $\mathbf{q}(\boldsymbol{\mu})$ in a latent space to a solution $\mathbf{u}(\boldsymbol{\mu})$ in the full space, and $n\ll N$. 

Correspondingly, we introduce its encoder:
\begin{equation}
\begin{aligned}
E_u: \ & \mathbb{R}^{N} \rightarrow \mathbb{R}^n, \
& \mathbf{u}(\boldsymbol{\mu}) \mapsto \mathbf{q}(\boldsymbol{\mu}).
\end{aligned}
\end{equation}
When Eq.~\eqref{eq: Genral Decoder} is substituted into Eq.~\eqref{eq: residual discrete}, we obtain a residual characterized by:
\begin{equation}
\mathbf{R}(\mathbf{u}(\boldsymbol{\mu});\boldsymbol{\mu}) =  \mathbf{R}(D_u(\mathbf{q}(\boldsymbol{\mu}));\boldsymbol{\mu}).
    \label{eq: Discrete residual (decoder)}
\end{equation}

Finding the best reduced solution $\boldsymbol{q}$ involves solving the following minimization problem:
\begin{equation}
    \min_{\mathbf{q}(\boldsymbol{\mu}) \in \mathbb{R}^n} \frac{1}{2}\left\lVert \mathbf{R}(D_u(\mathbf{q}(\boldsymbol{\mu}));\boldsymbol{\mu}) \right\rVert_{\mathbf{G}}^2  \ .
    \label{Minimization of the residual}
\end{equation}
Particular choices of the norm-defining matrix $\mathbf{G} \in \mathbb{R}^{N\times N}$ lead to different methods. In particular setting $\boldsymbol{G} := \boldsymbol{J}^{-T}$ leads to the Galerkin method, which we utilize in this work. For the derivation of the Galerkin method and other methods by particular choices of $\boldsymbol{G}$, the interested reader is directed to \cite{Ares2023}.

In this way, given an initial guess for the ROM solution, symbolically represented as $\mathbf{u}^0(\boldsymbol{\mu}) \in \mathbb{R}^{N}$, the ROM solution can be obtained using the Galerkin method by solving iteratively the following system of equations
\begin{subequations}
\label{eq:fixed-point_iteration_method_nmrom}
\begin{align}
    \label{eq:fixed-point_iteration_method_nmrom_a}
    \left( \frac{\partial D_u(\mathbf{q}^k(\boldsymbol{\mu}))}{\partial \mathbf{q}} \right)^T \mathbf{J} (\mathbf{u}^k(\boldsymbol{\mu}) ; \boldsymbol{\mu}) \frac{\partial D_u(\mathbf{q}^k(\boldsymbol{\mu}))}{\partial \mathbf{q}}  \delta\mathbf{q}^k(\boldsymbol{\mu})&= -\left( \frac{\partial D_u(\mathbf{q}^k(\boldsymbol{\mu}))}{\partial \mathbf{q}} \right)^T\mathbf{R}(\mathbf{u}^k(\boldsymbol{\mu}) ; \boldsymbol{\mu}) \\
    \mathbf{q}^{k+1}(\boldsymbol{\mu}) & = \mathbf{q}^{k}(\boldsymbol{\mu}) + \delta\mathbf{q}^k(\boldsymbol{\mu}) \\
    \mathbf{q}^0(\boldsymbol{\mu}) &= E_u(\mathbf{u}^0(\boldsymbol{\mu}))
\end{align}
\end{subequations}

\subsection{Neural Network-Augmented Projection-Based ROM (PROM-ANN)} \label{subsec: neural-network-pod}

In this work, we focus on the PROM-ANN proposed by Barnett et al.\cite{BARNETT2023112420}. This methodology constructs a nonlinear approximation manifold by augmenting a traditional linear reduced-order basis ROB with a nonlinear correction obtained via an artificial neural network (ANN). It builds upon earlier developments in quadratic approximation manifolds\cite{barnett2022quadratic}, addressing limitations inherent to purely linear ROM approximations, especially regarding problems exhibiting slowly decaying Kolmogorov n-width.

\subsubsection{Construction of the Nonlinear Approximation Manifold}

The PROM-ANN methodology constructs the solution manifold by leveraging a low-dimensional primary basis and enhancing it with a nonlinear mapping learned from data. Specifically, POD is employed by factorizing a snapshot matrix $\mathbf{S}_u\in\mathbb{R}^{N\times m}$ (containing $m$ high-fidelity solutions) via a truncated singular value decomposition:

\begin{equation} \mathbf{S}_u = \mathbf{U}\boldsymbol{\Sigma}\mathbf{V}^T, \end{equation}

\noindent where $\mathbf{U} \in \mathbb{R}^{N \times (n+\bar{n})}$ contains the left singular vectors, and $\boldsymbol{\Sigma}$ is the diagonal matrix of singular values $\sigma_i$, ordered by magnitude. From this decomposition, we construct the primary ROB $\boldsymbol{\Phi}\in\mathbb{R}^{N\times n}$ using the first $n$ dominant modes, and a secondary ROB $\bar{\boldsymbol{\Phi}}\in\mathbb{R}^{N\times \bar{n}}$ using the next $\bar{n}$ subdominant modes. Given a full-order solution vector $\mathbf{u}^*(\boldsymbol{\mu}) \in \mathbb{R}^{N}$, we approximate it with a nonlinear manifold representation defined as follows:

\begin{subequations}
\label{eq:original-prom-ann-encod-decod} 
\begin{align}
\mathbf{u}^*(\boldsymbol{\mu}) \approx \mathbf{u}(\boldsymbol{\mu}) &= D_u(\mathbf{q}(\boldsymbol{\mu})) = \mathbf{u}_\text{ref} + \PhiInf\mathbf{q}(\boldsymbol{\mu}) + \PhiSup\mathcal{N}(\mathbf{q}(\boldsymbol{\mu}); \boldsymbol{\Theta}) \\
\mathbf{q}(\boldsymbol{\mu}) &= E_u(\mathbf{u}^*(\boldsymbol{\mu})) = \PhiInf^T(\mathbf{u}^*(\boldsymbol{\mu})-\mathbf{u}_\text{ref}),
\end{align}
\end{subequations}

\noindent where $\mathbf{u}_\text{ref} \in \mathbb{R}^{N}$ is a suitable reference solution, typically chosen as the mean of the snapshots, $\mathbf{q}(\boldsymbol{\mu}) \in \mathbb{R}^{n}$ are the primary reduced coordinates, encoding the dominant features of the solution, $\mathcal{N}(\mathbf{q}; \boldsymbol{\Theta}): \mathbb{R}^{n} \rightarrow \mathbb{R}^{\bar{n}}$ is a neural network with parameters $\boldsymbol{\Theta}$, creating a nonlinear mapping from the primary to the secondary reduced coordinates, thus providing a nonlinear correction to the linear subspace approximation.

The ANN acts as a nonlinear map, capturing correlations between dominant (primary) and subdominant (secondary) mode coefficients. Rather than explicitly reconstructing truncated information, the ANN provides a nonlinear relationship that significantly improves the compactness and accuracy of the reduced-order representation. 

To solve the reduced-order system defined in Eq.~\eqref{eq:fixed-point_iteration_method_nmrom}, it is necessary to explicitly compute the derivative of the decoder function with respect to the primary reduced coordinates. For the PROM-ANN architecture defined in Eq.~\eqref{eq:original-prom-ann-encod-decod}, this derivative reads:

\begin{equation} \frac{\partial D_u(\mathbf{q}(\boldsymbol{\mu}))}{\partial \mathbf{q}} = \boldsymbol{\Phi} + \bar{\boldsymbol{\Phi}} \frac{\partial \mathcal{N}(\mathbf{q}(\boldsymbol{\mu}); \boldsymbol{\Theta})}{\partial \mathbf{q}}. \label{eq: decoder_derivative} \end{equation}

The derivative of the neural network term $\frac{\partial \mathcal{N}(\mathbf{q}(\boldsymbol{\mu}); \boldsymbol{\Theta})}{\partial \mathbf{q}}$ can be computed efficiently using automatic differentiation frameworks typically employed in neural network training.

\subsubsection{Training the Neural Network}

The ANN is trained using precomputed solution snapshots from the high-fidelity model. For each training snapshot indexed by $j = 1, \dots, m$, we first compute the primary and secondary reduced coordinates by projection onto their respective ROBs:

\begin{subequations}
\begin{align}
    \mathbf{q}_j(\boldsymbol{\mu}_j) &= \boldsymbol{\Phi}^T (\mathbf{u}^*_j(\boldsymbol{\mu}_j) - \mathbf{u}_\text{ref}), \\
    \bar{\mathbf{q}}_j(\boldsymbol{\mu}_j) &= \bar{\boldsymbol{\Phi}}^T (\mathbf{u}^*_j(\boldsymbol{\mu}_j) - \mathbf{u}_\text{ref}).
\end{align}
\end{subequations}

The original PROM-ANN approach~\cite{BARNETT2023112420} defines the training loss function directly on the discrepancy between the secondary coordinates $\bar{\mathbf{q}}_j$ and the ANN prediction $\mathcal{N}(\mathbf{q}_j; \boldsymbol{\Theta})$:

\begin{equation}
\label{eq:original_pod-ann_loss}
\mathcal{L}_\text{PROM-ANN} = \frac{1}{m} \sum_{j=1}^{m} \| \bar{\mathbf{q}}_j(\boldsymbol{\mu}_j) - \mathcal{N}(\mathbf{q}_j(\boldsymbol{\mu}_j); \boldsymbol{\Theta}) \|^2.
\end{equation}

While very computationally efficient, this approach lacks a proper normalization strategy in order to guarantee that the neural network can be effectively trained. We will study this issue in Section \ref{sec:modification_to_PROM-ANN_architecture}.

% In order to incorporate the physics, we extend the loss formulation to directly operate on the full-order solution vector as follows:

% \begin{equation} 
% \boldsymbol{\Theta}^* = \arg\min_{\boldsymbol{\Theta}} \underbrace{\frac{1}{mN} \sum_{j=1}^{m} \| \mathbf{u}_\text{ref} + \boldsymbol{\Phi}\mathbf{q}_j(\boldsymbol{\mu}_j) + \bar{\boldsymbol{\Phi}}\mathcal{N}(\mathbf{q}_j(\boldsymbol{\mu}_j); \boldsymbol{\Theta}) - \mathbf{u}^*_j(\boldsymbol{\mu}_j) \|^2}_{\mathcal{L_{d}}}, \label{eq: ann-loss-function-u} \end{equation}
% or
% \begin{equation} 
% \boldsymbol{\Theta}^* = \arg\min_{\boldsymbol{\Theta}} \underbrace{\frac{1}{mN} \sum_{j=1}^{m} \| D_u(E_u(\mathbf{u}_j(\boldsymbol{\mu}_j));\boldsymbol{\Theta}) - \mathbf{u}^*_j(\boldsymbol{\mu}_j) \|^2}_{\mathcal{L_{d}}}, \label{eq: ann-loss-function-decoder} \end{equation}
% thus explicitly minimizing the discrepancy at the level of the full-order solution. This modification to the loss function provides a natural pathway toward embedding discrete PDE residuals, which we explore in the subsequent sections.
\section{Discrete PINN-like Loss}
\label{sec:discrete_pinn}

Our non-linear ROM architecture aims to incorporate the physics of the problem into the approximation manifold construction. The equivalent effect is accomplished in traditional PINNs by auto-differentiating the neural network's output with regard to a given point in continuous space and time, using the strong form of the PDE system. In contrast, a discrete approach like ours, requires a numerical approximation by discretization of the PDE, which can be done using a variety of techniques such as the finite element, or finite volume methods. The integration of these discretized residuals into neural networks places our approach within the category of informed machine learning, as detailed by \cite{VonRueden2023}.

The proposal to use discrete approximation for substituting the auto-differentiation in PINNs for residual minimization is introduced almost simultaneously in \cite{Meethal2023} and \cite{Thang2023}. Both of these approaches introduce NN architectures for solving forward problems of linear, steady-state simulation and \cite{Meethal2023} additionally presents a model for backwards problems of the same nature. Their conceptualization doesn't differ too much from classical PINNs in terms of inputs, outputs, and loss definition: they take the parameters vector for the desired simulation as input, and return the results of the nodal variables of the system as output. The spatial information that is typically an input in PINNs is intrinsically defined in the FEM solver at the time of the residual computation.

In terms of the training loss, both approaches aim to minimize the mean squared L2-norm of the residual, as determined by the FEM solver. This minimization considers the predicted snapshot (nodal variables) and the relevant simulation parameters:
\begin{equation}
\label{eq:pinn_residual_loss}
    \mathcal{L}_{\mathbf{R}} = \frac{1}{m N} \sum_{j=1}^{m} \norm{ \mathbf{R}(\mathcal{N}(\boldsymbol{\mu}_j ; \boldsymbol{\Theta}); \boldsymbol{\mu}_j)}^{2}  \ ,
\end{equation}
where $\boldsymbol{\Theta}$ are the trainable parameters of the neural network conforming the PINN, index $j$ specifies the different samples to be used within the batch, $m$ is the number of samples in the batch and $N$ is the number of degrees of freedom in our system (and therefore the size of the residual). An important remark about this loss is that it requires the removal of contributions in the residual $\mathbf{R}(\mathcal{N}(\boldsymbol{\mu}_j ; \boldsymbol{\Theta}); \boldsymbol{\mu}_j)$ at the degrees of freedom with Dirichlet conditions, otherwise the norm will not approach zero.

The exact implementation of this loss differs in both approaches. The one in \cite{Thang2023} is limited to static linear cases of structural mechanics and proposes a specific loss scaling strategy for these cases. They also propose an implementation strategy to perform the loss computation in batches. Meanwhile, the approach in \cite{Meethal2023} is more general, but still only applicable to linear PDEs. This limitation to linear cases significantly simplifies the loss implementation, as the residual becomes of the form $\mathbf{R}(\mathbf{u};\boldsymbol{\mu}) = \mathbf{K}(\boldsymbol{\mu}) \mathbf{u} - \mathbf{F}(\boldsymbol{\mu})$. So both the stiffness matrix $\mathbf{K}$ and the vector of external contributions $\mathbf{F}$ are independent of the current snapshot. Accordingly, the loss becomes:
% \begin{equation}
% \frac{\partial\mathcal{L}_{\mathbf{R}}}{\partial\boldsymbol{\Theta}} = \frac{2}{m N} \sum_{j=1}^{m} (\mathbf{K}(\boldsymbol{\mu}_j) \mathcal{N}(\boldsymbol{\mu}_j ; \boldsymbol{\Theta}) - \mathbf{F}(\boldsymbol{\mu}_j) ) \mathbf{K}(\boldsymbol{\mu}_j) \frac{\partial \mathcal{N}(\boldsymbol{\mu}_j ; \boldsymbol{\Theta})}{\partial \boldsymbol{\Theta}}.
% \end{equation}
\begin{equation}
\mathcal{L}_{\mathbf{R}} = \frac{1}{m N} \sum_{j=1}^{m} \norm{\mathbf{K}(\boldsymbol{\mu}_j) \mathcal{N}(\boldsymbol{\mu}_j ; \boldsymbol{\Theta}) - \mathbf{F}(\boldsymbol{\mu}_j)}^{2}
\end{equation}
where one could have precomputed all $\mathbf{K}(\boldsymbol{\mu}_j)$ and $\mathbf{F}(\boldsymbol{\mu}_j)$, thus allowing the whole optimization to be done via auto-differentiation, without calling the FEM software.

Even the approaches in more recent papers like \cite{yamazaki_finite_2025} and \cite{sunil_fe-pinns_2024} are still designed only for linear problems, with the difference being their specific use cases, the architectures they use for the surrogate model and the loss being just the norm of the residual, instead of the square of it.

In contrast, our approach differs in two key aspects: (1) the loss function is formulated in a parameter-agnostic manner, and (2) it is designed to handle both linear and non-linear cases, thereby enabling a much broader range of applications. As with PINNs, our method also allows for a combination of physics-based and data-driven losses. The following subsections discuss these three developments.

\subsection{Parameter-agnostic loss}

One aspect in which our methodology diverges significantly from the original idea of PINNs is that we are not training a self-contained surrogate model (typically in the form of a single neural network $\mathcal{N}(\boldsymbol{\mu}_j)$) but instead, the approximation manifold defined by a trainable encoder-decoder pair $D_u(E_u(\mathbf{u}^*_{j}))$. As we can see, the latter case does not take the simulation parameters $\boldsymbol{\mu}_j$ as an input, so it is agnostic to them. This is because the FEM-based ROM software itself will be in charge of conditioning the simulation with those parameters and then finding the most appropriate solution within our latent space.

But not only that. Traditionally, for a loss like the one in Eq.~\eqref{eq:pinn_residual_loss}, we need to specify the exact parameters $\boldsymbol{\mu}_j$ for the solution we are seeking, as that is the one that will yield a residual of zero and therefore enable the learning by minimization. Instead, we design a loss function in which the parameters applied in the FEM software can be arbitrary. In this new loss we are not merely minimizing the properly parameterized residual of the predicted quantity $D_u(E_u(\mathbf{u}^*_{j}))$, but the difference between this quantity and the residual of the target solution $\mathbf{u}^*_{j}$, both with constant parameter $\boldsymbol{\mu}$:

\begin{equation}
\label{eq:our_residual_loss}
    \mathcal{L}_{\mathbf{R}} =  \frac{1}{m N} \sum_{j=1}^{m} \norm{ \mathbf{R}(D_u(E_u(\mathbf{u}^*_{j}));  \boldsymbol{\mu} ) - \mathbf{R}(\mathbf{u}^*_{j}; \boldsymbol{\mu}) }^{2}  \ .
\end{equation}

In this case, the trainable parameters $\boldsymbol{\Theta}$ are contained within the encoder and/or decoder of our ROM, and their specific form depends on the chosen architecture. Strictly speaking, we should write $D_u(E_u(\mathbf{u}^*_{j}; \boldsymbol{\Theta}); \boldsymbol{\Theta})$. However, for clarity and conciseness, we will omit $\boldsymbol{\Theta}$ from the notation in what follows.

Our approach aims to minimize the discrepancy in residual behavior when it is non-zero, ensuring the ROM approximation manifold captures not only the converged solution but also the behavior of the residual outside of convergence situations. In our approach, the decoder will be readily integrated into the Newton iterative procedure. Thus, we aim for it to enhance its residual representation near the convergence space, aiding in achieving an optimal converged solution. On the same line, we no longer have the restriction that we mentioned for Eq.~\eqref{eq:pinn_residual_loss} on the components of the residual associated to Dirichlet conditions. Meaning that we can leave those components in order to learn them too. The main caveat of using this approach is that it still requires the original HFM's snapshots samples $\mathbf{u}^*_{j}$ even when training via the residual. In our specific case, the encoder-decoder architecture of choice will need these snapshots regardless, so it is not a major inconvenient.

Regarding boundary conditions, it is important to consider that Dirichlet conditions must be imposed strongly in the solutions vector whenever we compute the residual. That means that whatever architecture we use as encoder-decoder pair, it needs to operate on and predict only the degrees of freedom unaffected by Dirichlet conditions. Then, the fixed ones are given their corresponding value forcefully before computing the residual and Jacobian.

\subsection{In-training integration of FEM software}
\label{subsec:In-training integration of FEM software}

The obstacle impeding going from linear to non-linear cases in other FEM-based residual losses \cite{Meethal2023,Thang2023}  is not a theoretical or mathematical difficulty in generating the residuals themselves or their derivatives. We have extensive FEM software at our disposal that already does precisely this. The difficulty is in dynamically integrating this information in an effective way during the training of the decoder.

Let us study the loss we proposed in Eq.~\eqref{eq:our_residual_loss} in more detail to identify our needs during training. The loss itself needs the values for two residuals: $\mathbf{R}(D_u(E_u(\mathbf{u}^*_{j}));  \boldsymbol{\mu} )$ and $\mathbf{R}(\mathbf{u}^*_{j}; \boldsymbol{\mu})$. The latter can be pre-computed, as it will be constant for sample $j$ during the whole training. The first one, however, has to be computed within the FEM software with an updated snapshot value at each training step.

Now, the actual training of the model happens during the backpropagation of the loss, which necessitates calculating the gradient of the loss with respect to $\boldsymbol{\Theta}$. This is typically done automatically thanks to the auto-differentiation capabilities of deep learning frameworks. However, as we take part of the loss computation onto external software, we are unable to do this and we have to code the gradient manually. The derivative of the loss function $\mathcal{L}_{\mathbf{R}}$ can be decomposed using the chain rule:
\begin{equation}
\label{eq:our_residual_loss_gradient}
\frac{\partial\mathcal{L}_{\mathbf{R}}}{\partial\boldsymbol{\Theta}} = \frac{2}{m N} \sum_{j=1}^{m} \left( \mathbf{R}(\mathbf{u}_{j}; \boldsymbol{\mu}) - \mathbf{R}(\mathbf{u}^*_{j}; \boldsymbol{\mu}) \right)^T \frac{\partial \mathbf{R}(\mathbf{u}_{j}; \boldsymbol{\mu})}{\partial \mathbf{u}_{j}} \frac{\partial \mathbf{u}_{j}}{\partial \boldsymbol{\Theta}}.
\end{equation}
Where $\mathbf{u}_{j} = D_u(E_u(\mathbf{u}^*_{j}))$ is the current prediction for the nodal values of the solution.

The factor \(\frac{\partial \mathbf{R}(\mathbf{u}_{j}; \boldsymbol{\mu})}{\partial \mathbf{u}_{j}}\) is the Jacobian matrix of the residual function with respect to the predicted solution vector \(\mathbf{u}_{j}\). This Jacobian is computed and assembled entirely in the FEM software, given the current snapshot and arbitrary simulation parameter. Within the context of FEM, it is expressed as:

\begin{equation}
    \mathbf{J}_F(\mathbf{u}_{j}; \boldsymbol{\mu}) = \frac{\partial \mathbf{R}(\mathbf{u}_{j}; \boldsymbol{\mu})}{\partial \mathbf{u}_{j}} \ .
\end{equation}

The final factor $\frac{\partial \mathbf{u}_{j}}{\partial \boldsymbol{\Theta}}$ represents another Jacobian, this time of the predicted snapshot with respect to the trainable parameters. We can express it as:
\begin{equation}
    \mathbf{J}_{D}(\mathbf{u}_{j}; \boldsymbol{\mu}) = \frac{\partial\mathbf{u}_j}{\partial\mathbf{\Theta}} \ .
    \label{TF jacobian}
\end{equation} 
This latter derivative is applied only on the computations performed in the trainable encoder-decoder pair $D_u(E_u(\mathbf{u}^*_{j}))$. As such, this factor is self-contained within the deep learning framework and does not require additional external data. Such a Jacobian could be obtained by using the \texttt{tf.GradientTape.batch\_jacobian()} method in TensorFlow, or equivalent methodologies in other deep learning frameworks.

To compute the loss gradient as it is defined explicitly in Eq.~\eqref{eq:our_residual_loss_gradient}, we just need to collect the factors coming from the FEM software into the deep learning framework. However, this computation would be unnecessarily inefficient because of two main reasons described next. For each of these, we propose a corresponding implementation strategy:

\begin{enumerate}
    \item First, take into account that $\mathbf{J}_F(\mathbf{u}_{j}; \boldsymbol{\mu})$ is typically a high-dimensional and highly sparse matrix. So, moving it directly from one software to another could be costly, especially if it cannot be treated as a sparse structure in all computations. FEM software is typically designed to cope with these kinds of matrices and to perform sparse vector-matrix multiplications. Therefore, for each snapshot we precompute the quantities:
    \begin{equation}
    \begin{aligned}
    &\mathbf{e}_{\mathbf{R},j} \coloneqq \left( \mathbf{R}(\mathbf{u}_{j}; \boldsymbol{\mu}) - \mathbf{R}(\mathbf{u}^*_{j}; \boldsymbol{\mu})\right)^T\\
    &\mathbf{v}_{\mathbf{R},j} \coloneqq \mathbf{e}_{\mathbf{R},j} \mathbf{J}_F(\mathbf{u}_{j}; \boldsymbol{\mu})
    \end{aligned}
    \end{equation}
    within the FEM software. Now both of these are vectorial quantities, making the transfer to the deep learning framework more efficient. $\mathbf{e}_{\mathbf{R},j}$ is the error used to compute the loss, as $\mathcal{L}_{\mathbf{R}} = \frac{1}{m N} \sum_{j=1}^{m} \norm{\mathbf{e}_{\mathbf{R},j}}^{2}$. And $\mathbf{v}_{\mathbf{R},j}$ will be used to compute the gradient of the loss.
    \item Then, the main bottleneck comes from needing to compute $m$ full Jacobian matrices $\{\mathbf{J}_{D}(\mathbf{u}_{j}; \boldsymbol{\mu})\}_{j=1}^m$ via auto-differentiation, instead of just the gradient of a single scalar $\mathcal{L}_{\mathbf{R}}$. Avoiding this is straightforward when reformulating the gradient of the loss with vectors $\mathbf{v}_{\mathbf{R},j}$. We realize that, to the deep learning framework, externally-computed $\mathbf{v}_{\mathbf{R},j}$ is no longer seen as a function depending on the model parameters $\boldsymbol{\Theta}$, but as a constant instead. This means that we can treat it as such during the computation and get:
    \begin{equation}
    \frac{\partial\mathcal{L}_{\mathbf{R}}}{\partial\boldsymbol{\Theta}} = \frac{2}{m N} \sum_{j=1}^{m} \mathbf{v}_{\mathbf{R},j} \frac{\partial \mathbf{u}_{j}}{\partial \boldsymbol{\Theta}} = \frac{\partial }{\partial \boldsymbol{\Theta}} \left( \frac{2}{m N} \sum_{j=1}^{m} \mathbf{v}_{\mathbf{R},j} \mathbf{u}_{j} \right)
    \end{equation}
    Therefore, we compute a single gradient of a scalar value for the whole batch, via auto-differentiation.
\end{enumerate}

Particularly in our case, we take KratosMultiphysics as our choice of FEM software. As it is open-source, it enables us to develop all the custom functionalities needed to take a given set of snapshots $\{\mathbf{u}_{j}\}_{j=1}^m$ (those of the current batch), and for each one apply it as the current solution to then output $\mathbf{e}_{\mathbf{R},j}$ and $\mathbf{v}_{\mathbf{R},j}$.

\subsection{Data-based loss term}
\label{subsec:data-basesd_loss_term}

Even when training with the residual loss, it might be beneficial to introduce a data-related loss simultaneously. This is common in traditional PINNs in two ways: specifically on the boundary and initial conditions, to try to enforce Dirichlet conditions \cite{raissi_physics_2017} (unless a special approach is used to establish them strongly \cite{SUN2020112732}), and on collocation points sampled inside the simulation space, as a regularization that can help the overall loss converge faster \cite{cuomo2022scientific}.

In our case the Dirichlet conditions are enforced strongly and managed by the FEM software for their effect on the residual. But we could still introduce such a loss term for regularization purposes. Thus, the total loss function would be defined by two components:
\begin{equation}
\label{eq:our_total_loss}
\mathcal{L} = \omega_{\mathbf{R}} \mathcal{L}_{\mathbf{R}} + \omega_{d}\mathcal{L}_{d} \ .
\end{equation}
\noindent where $\omega_{\mathbf{R}}$ and $\omega_{d}$ are tunable hyper-parameters to balance both terms as needed, and the data-related loss $\mathcal{L}_{d}$ is defined simply as the mean squared error between the currently predicted snapshot $\mathbf{u}_{j} = D_u(E_u(\mathbf{u}^*_{j}))$ and the ground-truth one $\mathbf{u}^*_{j}$:
\begin{equation}
\label{eq:our_data_loss}
\mathcal{L}_{d} = \frac{1}{m N} \sum_{j=1}^{m} \norm{ \mathbf{u}_{j}  - \mathbf{u}^*_{j} }^{2} \ .
\end{equation}

\noindent Then, the implementation of the total loss $\mathcal{L}$ and its gradient starts by defining the following vectors as constants:

\begin{equation}
\begin{aligned}
\label{eq:constant_vectors_definitions}
\mathbf{e}_{d,j} = \mathbf{v}_{d,j} &\coloneqq (\mathbf{u}_{j} - \mathbf{u}^*_{j})^T \ ,\\
\mathbf{e}_{\mathbf{R},j} &\coloneqq \left( \mathbf{R}(\mathbf{u}_{j}; \boldsymbol{\mu}) - \mathbf{R}(\mathbf{u}^*_{j}; \boldsymbol{\mu})\right)^T \ , \\
\mathbf{v}_{\mathbf{R},j} &\coloneqq \mathbf{e}_{\mathbf{R},j} \mathbf{J}_F(\mathbf{u}_{j}; \boldsymbol{\mu}) \ ,
\end{aligned}
\end{equation}

\noindent and then using these to compute the loss as
\begin{equation}
\mathcal{L} = \frac{1}{m N} \sum_{j=1}^{m} \left ( \omega_{\mathbf{R}} \norm{\mathbf{e}_{\mathbf{R},j}}^2 + \omega_{d} \norm{\mathbf{e}_{d,j}}^2 \right )
\end{equation}

\noindent And finally its gradient via auto-differentiation:

\begin{equation}
\frac{\partial\mathcal{L}}{\partial\boldsymbol{\Theta}} = \frac{\partial }{\partial \boldsymbol{\Theta}} \left( \frac{2}{m N} \sum_{j=1}^{m} \left( \omega_{\mathbf{R}} \mathbf{v}_{\mathbf{R},j} + \omega_{d} \mathbf{v}_{d,j} \right ) \mathbf{u}_{j} \right) \ .
\end{equation}

\section{Modifications to the PROM-ANN architecture}
\label{sec:modification_to_PROM-ANN_architecture}

We adopt the PROM-ANN architecture from \cite{BARNETT2023112420} as the foundation for developing our physics-informed ROM model, which introduces the use of neural networks while ensuring scalability and avoiding the requirement of a structured mesh for the underlying simulation. The architecture was introduced in Section~\ref{subsec: neural-network-pod}, where we noted that it can be interpreted as a non-linear extension of classical POD that introduces additional effective modes without increasing the size of the latent space.

However, we find that the architecture, as originally proposed, is difficult to train---at least in our case, where the underlying problem exhibits a rapidly decaying singular value spectrum. To address this, we first revise both the architecture and the data-driven loss formulation to improve training capability. Later, in the following section, we will then incorporate our residual-based loss into the framework.

The modifications we propose in this section are as follows: (1) scaling of reduced coefficients, (2) changing the data-based loss for one that takes the full snapshot into account and (3) properly scaling the loss

\subsection{Scaling of reduced coefficients}

The main concern for us from the original architecture is that there is no normalization or scaling of the reduced coefficients prior to using them in the neural network. Ideally, neural networks are designed to operate on inputs that are approximately independent and identically distributed (i.i.d.), as this assumption facilitates more effective learning and optimization \cite{montavon2012neural, goodfellow2016deep}. This can rarely be guaranteed, but it is still common practice to perform some sort of normalization or scaling on the inputs so that they all operate in a similar range of values. Otherwise, serious issues may arise during training, mainly because of some of the inputs being ignored.

In the case of PROM-ANN, the inputs and outputs of the neural network are the coefficients for a given snapshot (remember from Section \ref{subsec: neural-network-pod} that we  train a network such that $\mathcal{N}(\boldsymbol{\Phi}^T \mathbf{u}^*_j) \approx \bar{\boldsymbol{\Phi}}^T \mathbf{u}^*_j$). These coefficients $\mathbf{q}^*_j=\boldsymbol{\Phi}^T \mathbf{u}^*_j$ and $\bar{\mathbf{q}}^*_j=\bar{\boldsymbol{\Phi}}^T \mathbf{u}^*_j$ will scale similarly to the singular value decay observed when performing SVD on the training data. Essentially, for a fast-decaying singular value profile, each coefficient will result in a considerably smaller range of coefficients than the previous one, causing the neural network to  learn only from a few of the first ones.

In order to correct this issue, we modify the architecture of the encoder and the decoder themselves to include a pair of scaling matrices\footnote{This formulation makes the assumption that the problem is homogenous, which is true for our particular use-case. Still, it can be applied to non-homogenous cases. In such cases the user should remove the components affected by Dirichlet conditions from all snapshots and proceed as stated. Then, at the time of running the online simulation, the fixed components of the solution can be imposed strongly, as they are known.}:
\begin{equation}
\label{eq:encoder_decoder_ours}
    \begin{aligned}
        &E_u(\mathbf{u^*}) = \qInf
        = \boldsymbol{\XiInf}^{-1}\boldsymbol{\PhiInf}^T \mathbf{u^*}\\
        &D_u(\qInf) = \PhiInf \boldsymbol{\XiInf} \qInf + \PhiSup \XiSup \mathcal{N}(\qInf)
    \end{aligned}
\end{equation}
where both the projection matrices $\PhiInf$, $\PhiSup$ and the scaling matrices $\XiInf$, $\XiSup$ come from the SVD decomposition of the snapshots matrix:
\begin{equation}
    \begin{aligned}
        &SVD\left(\mathbf{S_u}\right)=\mathbf{U}\Sigma\mathbf{V}^T \\
        &\PhiInf=[\mathbf{U}_1|...|\mathbf{U}_n], \quad 
        \PhiSup=[\mathbf{U}_{n+1}|...|\mathbf{U}_{\bar{n}}] \\
        &\XiInf=\frac{1}{\sqrt{M}}\begin{bmatrix}
                \sigma_1 & &  &  \\
                 & \sigma_2 & &  \\
                 & & ... & \\
                 & &  & \sigma_n
            \end{bmatrix}, \quad
        \XiSup=\frac{1}{\sqrt{M}}\begin{bmatrix}
                \sigma_{n+1} & &  &  \\
                 & \sigma_{n+2} & &  \\
                 & & ... & \\
                 & &  & \sigma_{\bar{n}}
            \end{bmatrix} \\
    \end{aligned}
\end{equation}
and $\sigma_i$ are the singular values found in the diagonal of the $\boldsymbol{\Sigma}$ matrix. Finally, $M$ is the total number of samples in $\mathbf{S_u}$.

This procedure effectively re-scales each coefficient to have a roughly equivalent range. In the ideal case where all rows of the training snapshot matrix $\mathbf{S}_u$ have zero mean, multiplying by the matrix $\XiSup^{-1}$ scales the quantity $\mathbf{Q} = \XiSup^{-1} \mathbf{U}^T \mathbf{S}_u$ such that the covariance matrix between its rows becomes the identity. In other words, the modes in $\mathbf{Q}$ are uncorrelated—already ensured by the projection onto $\boldsymbol{\PhiInf}$—and have unit variance.

In our case, the condition that all row means in $\mathbf{S}_u$ are exactly zero does not hold. Nevertheless, as we will show in Section~\ref{sec: usecase and evaluation methodology}, the resulting scaling still leads to reduced coefficients with similar magnitudes in our particular use case.

We intentionally avoid enforcing zero-mean values, as we do not wish to apply any offset to the projected values. This design choice is crucial because it enables a neural network architecture without biases, thereby ensuring that $D_u(E_u(\mathbf{0})) = \mathbf{0}$ holds. Additionally, the proposed scaling procedure preserves the orthogonality of the projection matrices, so we still have $E_u(D_u(\mathbf{q})) = \mathbf{q}$.

Finally, no additional computation is required to obtain the scaling factors, since the SVD is already performed as part of the original methodology. In this sense, the approach is highly efficient.

\subsection{Corrected data-based loss}

Now we have an architecture that provides appropriately-scaled features for the neural network. But at the same time, this scaling may not be the best if we want to apply the training loss as in the original paper\cite{BARNETT2023112420}. That is, measuring the error on the predicted $\bar{\mathbf{q}}_j$ coefficients themselves. This loss would translate like this to our architecture:
\begin{equation}
\label{eq:initial_modified_pod-ann_loss}
    \mathcal{L}_{d}' = \frac{1}{mN} \sum_{j=1}^{m} 
    \norm{\mathcal{N}(\qInf^*_j) - \XiSup^{-1}\PhiSup^T \mathbf{u}^*_j}^2 = \frac{1}{mN} \sum_{l=1}^{m} 
    \norm{\mathcal{N}(\XiInf^{-1} \PhiInf^T \mathbf{u}^*_j) - \XiSup^{-1}\PhiSup^T \mathbf{u}^*_j}^2
\end{equation}
The rationale behind dividing by $N$ is to ensure consistency with the loss formulations introduced in Section~\ref{sec:discrete_pinn}.

Such a loss will give the same importance to all of the output features, or reduced coefficients in this case. We know from the nature of POD that this is not desirable, as lower modes should be given more importance than higher ones. In fact what we have to do is to reverse, within the loss, the scaling that we applied to the reduced coefficients. We can do so by applying $\bar{\boldsymbol{\Xi}}$ on the contents of the norm in Eq.~\eqref{eq:initial_modified_pod-ann_loss}:
\begin{equation}
\label{eq:our_prom-ann_data_loss_reduced}
    \mathcal{L}_{d,\text{compact}} = \frac{1}{mN} \sum_{l=1}^{m} 
    \norm{\XiSup \left(\mathcal{N}(\qInf^*) - \XiSup^{-1}\PhiSup^T \mathbf{u}^* \right)}^2
\end{equation}

But we could also achieve the same effect by just enforcing the whole reconstructed full-order snapshot to approximate the ground-truth one. That is, using exactly the same data-based loss that we had defined in Section \ref{subsec:data-basesd_loss_term}:
\begin{equation}
\label{eq:our_prom-ann_data_loss_unscaled}
\mathcal{L}_d = \frac{1}{m N} \sum_{j=1}^{m} \norm{ \mathbf{u}_{j}  - \mathbf{u}^*_{j} }^{2} = \frac{1}{m N} \sum_{j=1}^{m} \norm{ \PhiInf \XiInf \qInf^* + \PhiSup \XiSup \mathcal{N}(\qInf^*)  - \mathbf{u}^*_{j} }^{2}
\end{equation}

This second loss is more intuitive, as we make it explicit to achieve the final goal for our ROM approximation manifold: learning the full snapshot itself. However, it is not difficult to prove that both Eq.~\eqref{eq:our_prom-ann_data_loss_reduced} and Eq.~\eqref{eq:our_prom-ann_data_loss_unscaled} are equivalent up to a constant offset. That is, assuming that all snapshots $\{\mathbf{u}^*_j\}_{j=1}^m$ were included in the set $\mathbf{S_u}$ to which we performed the SVD.

\begin{proof}
By developing from $\mathcal{L}_d$:
\begin{equation*}
\mathcal{L}_{d} = \frac{1}{m N} \sum_{j=1}^{m} \norm{ \mathbf{u}_{j}  - \mathbf{u}^*_{j} }^{2} = \frac{1}{m N} \sum_{j=1}^{m} \norm{ \PhiInf \XiInf \qInf^* + \PhiSup \XiSup \mathcal{N}(\qInf^*)  - \mathbf{u}^*_{j} }^{2}
\end{equation*}
By developing $\mathbf{u}^*_{j} = (\PhiInf \XiInf \XiInf^{-1} \PhiInf^T + \PhiSup \XiSup \XiSup^{-1} \PhiSup^T + \hat{\boldsymbol{\Phi}} \hat{\boldsymbol{\Xi}} \hat{\boldsymbol{\Xi}}^{-1}\hat{\boldsymbol{\Phi}}^T) \mathbf{u}^*_j$. Where $\hat{\boldsymbol{\Phi}}$ is the orthonormal basis containing the singular vectors from $SVD(\mathbf{S_u})$ that were not included in neither $\PhiInf$ or $\PhiSup$, and $\hat{\boldsymbol{\Xi}}$ is analog but for the singular values matrix. Then applying it:
\begin{equation*}
\begin{aligned}
&\mathcal{L}_{d} = \frac{1}{m N} \sum_{j=1}^{m} \norm{ \PhiInf \XiInf \qInf^* + \PhiSup \XiSup \mathcal{N}(\qInf^*)  - \left(\PhiInf \XiInf \qInf^* + \PhiSup \XiSup \XiSup^{-1}\PhiSup^T \mathbf{u}^*_j + \hat{\boldsymbol{\Phi}} \hat{\boldsymbol{\Xi}} \hat{\boldsymbol{\Xi}}^{-1}\hat{\boldsymbol{\Phi}}^T \mathbf{u}^*_j \right)}^{2} \\
& = \frac{1}{m N} \sum_{j=1}^{m} \norm{ \PhiSup \XiSup \left( \mathcal{N}(\qInf^*) - \XiSup^{-1}\PhiSup^T \mathbf{u}^*_j\right) -  \hat{\boldsymbol{\Phi}}\hat{\boldsymbol{\Phi}}^T \mathbf{u}^*_j }^{2} \\
\end{aligned}
\end{equation*}
By acknowledging that $\PhiSup$ and $\hat{\boldsymbol{\Phi}}$ are orthogonal to each other:
\begin{equation*}
\mathcal{L}_{d} = \frac{1}{m N} \sum_{j=1}^{m} \left( \norm{ \PhiSup \XiSup \left( \mathcal{N}(\qInf^*) - \XiSup^{-1}\PhiSup^T \mathbf{u}^*_j\right)}^{2} + \norm{ \hat{\boldsymbol{\Phi}}\hat{\boldsymbol{\Phi}}^T \mathbf{u}^*_j }^{2} \right)
\end{equation*}
By acknowledging that the $L^2$ norm is invariant to right-multiplication of an orthonormal matrix and that the terms $\norm{ \hat{\boldsymbol{\Phi}}\hat{\boldsymbol{\Phi}}^T \mathbf{u}^*_j}$ do not depend on trainable parameters $\boldsymbol{\Theta}$:
\begin{equation*}
\mathcal{L}_{d} = \frac{1}{m N} \sum_{j=1}^{m} \norm{ \XiSup \left( \mathcal{N}(\qInf^*) - \XiSup^{-1}\PhiSup^T \mathbf{u}^*_j\right)}^{2} + C = \mathcal{L}_{d,\text{compact}} + C
\end{equation*}
\end{proof}

If the user does not intend on applying physics-informed training, then choosing $\mathcal{L}_{d,\text{compact}}$ would make most sense because of its reduced complexity. However, the fact that $\mathcal{L}_{d}$ takes the snapshots to the full space within the computation makes it most appropriate to be paired with a physics-based loss, which will require the full vector of nodal solutions anyways. In any of the two cases, the ideal would be to pre-compute $\qInf^*_j$ for all samples to improve efficiency.

\subsection{Scaling of the data-based loss}
\label{subsec:Scaling_of_the_data-based_loss}

As written in Eq.~\eqref{eq:our_prom-ann_data_loss_unscaled}, $\mathcal{L}_{d}$ would exhibit high variability depending on the number of modes included in the primary ROB. That is because, by including more modes in $\PhiInf$ while following their importance order, we are heavily limiting the possible error $\norm{ \mathbf{u}_{j}  - \mathbf{u}^*_{j} }^{2}$ that we can have \footnote{This is true for snapshots within the set $\mathbf{S_u}$, and should reflect in new samples if they are properly represented by the POD}.

In order to correct this issue, we propose to use a pre-computed scaling factor that will be applied globally to the loss:
\begin{equation}
\label{eq:our_prom-ann_data_loss}
\mathcal{L}_d = \frac{1}{m N} \frac{1}{e_{\text{POD},d}} \sum_{j=1}^{m} \norm{ \mathbf{u}_{j}  - \mathbf{u}^*_{j} }^{2} \ .
\end{equation}

With the new factor $e_{\text{POD},d}$ defined as:
\begin{equation}
e_{\text{POD},d} \coloneqq  \frac{1}{MN} \sum_{j=1}^{M} \norm{ \PhiInf\PhiInf^T\mathbf{u}^*_{j}  - \mathbf{u}^*_{j} }^{2} \ .
\end{equation}

This quantity corresponds to the mean squared reconstruction error of the standard POD approach when only the primary modes are retained. It is averaged over all the samples in the training set, not only those in the batch, in order to make it representative for any choice of samples during the training, and thus, only having to compute it once at the beginning.

This choice of scaling acts as a safeguard preventing the gradients from becoming too low, which could induce numerical instabilities during training, and prevents the weight updates from varying wildly in scale based on the chosen numbers of primary modes, for the same learning rate. Apart from this, it also gives a much more interpretable value to the loss, as it will become an indicator to how much better the current model is compared to just using POD for the same size of latent space. In this sense, it is helpful when troubleshooting, as a value over 1 would clearly indicate that we are losing accuracy compared to POD.

\subsection{Online phase of non-linear ROM}
\label{subsec:Online_phase_of non-linear_ROMÇ_ours}

In order to perform the online ROM simulation, we need to adapt the nonlinear iteration problem to the new architecture. Essentially, we substitute our decoder into the Galerkin-based non-linear iteration formulation for non-linear manifolds (as seen in Eq.~\eqref{eq:fixed-point_iteration_method_nmrom}). The result would be:
\begin{equation}
\begin{aligned}
\left(\PhiInf \XiInf + \PhiSup \XiSup \frac{\partial \mathcal{N}(\qInf^k)}{\partial \qInf}\right)^T \mathbf{J}(\mathbf{u}(\boldsymbol{\mu}) ; \boldsymbol{\mu}) \left(\PhiInf \XiInf + \PhiSup \XiSup \frac{\partial \mathcal{N}(\qInf^k)}{\partial \qInf}\right) \delta \textbf{q}^{k}(\boldsymbol{\mu}) = - \left(\PhiInf \XiInf + \PhiSup \XiSup \frac{\partial \mathcal{N}(\qInf^k)}{\partial \qInf}\right)^T \mathbf{R}(\mathbf{u}(\boldsymbol{\mu}) ; \boldsymbol{\mu}) \\
\textbf{q}^{k+1}(\boldsymbol{\mu}) = \textbf{q}^{k}(\boldsymbol{\mu}) + \delta\textbf{q}^k(\boldsymbol{\mu})
\end{aligned}
\end{equation}

Meaning that at each non-linear iteration we have to recompute $\PhiSup \XiSup \frac{\partial \mathcal{N}(\qInf^k)}{\partial \qInf}$.

The online ROM simulation is done entirely within the FEM software, KratosMultiphysics, as going back between this and the deep learning software would be unnecessarily expensive. Therefore, we made custom methods within KratosMultiphysics to compute both the forward pass of the neural network, $\mathcal{N}(\qInf)$, and the jacobian of its outputs with respect to the inputs, $\frac{\partial\mathcal{N}(\qInf)}{\partial \qInf}$. The other quantities are defined at initialization from the training results and are kept constant.

\section{Integration of physics-based loss into PROM-ANN}
\label{sec:integration_of_physics-based_loss_into_PROM-ANN}

Here we take the developments done in the two last sections and join them in order to finally obtain a method to perform physics-informed non-linear projection-based ROM. Essentially, we will take the total $\mathcal{L}$ loss developed in Section \ref{sec:discrete_pinn} and apply it on the PROM-ANN formulation developed in Section \ref{sec:modification_to_PROM-ANN_architecture}, while keeping in mind the adaptations to the loss that we also performed in this latter section.

The full discrete PINN-like loss $\mathcal{L}'$, as formulated in Section \ref{sec:discrete_pinn} is:

\begin{equation}
\label{eq:our_total_loss_unscaled}
\mathcal{L}' = \omega_{\mathbf{R}} \mathcal{L}'_{\mathbf{R}} + \omega_{d}\mathcal{L}'_{d} = \frac{\omega_{\mathbf{R}}}{m N} \sum_{j=1}^{m} \norm{ \mathbf{R}(\mathbf{u}_{j}; \boldsymbol{\mu})- \mathbf{R}(\mathbf{u}^*_{j}; \boldsymbol{\mu}) }^{2} + \frac{\omega_{d}}{m N} \sum_{j=1}^{m} \norm{ \mathbf{u}_{j}  - \mathbf{u}^*_{j} }^{2} \ .
\end{equation}

In Section \ref{subsec:Scaling_of_the_data-based_loss} we already determined that a full-snapshot data-based loss like $\mathcal{L}'_d$ in Eq.~\eqref{eq:our_total_loss_unscaled} is appropriate to use once scaled by factor $1/e_{\text{POD},d}$. So that is the only modification we have to do on $\mathcal{L}'_d$.

But now, in order to still be able to use the residual-based loss term $\mathcal{L}'_{\mathbf{R}}$ appropriately, we should also scale it in a similar way to $\mathcal{L}'_d$. We propose modifying it as follows:
\begin{equation}
\label{eq:our_prom-ann_residual_loss}
\mathcal{L}_{\mathbf{R}} = \frac{1}{m N} \frac{1}{e_{\text{POD},\mathbf{R}}} \sum_{j=1}^{m} \norm{ \mathbf{R}(\mathbf{u}_{j}; \boldsymbol{\mu})- \mathbf{R}(\mathbf{u}^*_{j}; \boldsymbol{\mu}) }^{2} \ .
\end{equation}

With $e_{\text{POD},\mathbf{R}}$ defined as the mean square error compared to the traditional POD, but this time in terms of the residual and not the snapshot itself:
\begin{equation}
e_{\text{POD},\mathbf{R}} = \frac{1}{MN} \sum_{j=1}^{M} \norm{ \mathbf{R}(\PhiInf\PhiInf^T\mathbf{u}^*_{j};  \boldsymbol{\mu} ) - \mathbf{R}(\mathbf{u}^*_{j}; \boldsymbol{\mu}) }^{2} \ . 
\end{equation}

By making the losses relative via $e_{\text{POD},\mathbf{R}}$ and $e_{\text{POD},d}$, the scale for both $\mathcal{L}_{\mathbf{R}}$ and $\mathcal{L}_{d}$ should be of similar scale. With this small modification, the full loss becomes:
\begin{equation}
\label{eq:our_total_loss}
\mathcal{L} = \omega_{\mathbf{R}} \mathcal{L}_{\mathbf{R}} + \omega_{d}\mathcal{L}_{d} = \frac{\omega_{\mathbf{R}}}{m N} \frac{1}{e_{\text{POD},\mathbf{R}}} \sum_{j=1}^{m} \norm{ \mathbf{R}(\mathbf{u}_{j}; \boldsymbol{\mu})- \mathbf{R}(\mathbf{u}^*_{j}; \boldsymbol{\mu}) }^{2} + \frac{\omega_{d}}{m N} \frac{1}{e_{\text{POD},d}}\sum_{j=1}^{m} \norm{ \mathbf{u}_{j}  - \mathbf{u}^*_{j} }^{2} \ .
\end{equation}

And the practical implementation of the loss itself and its gradient is:
\begin{equation}
\begin{aligned}
\label{eq:our_total_loss_and_gradient_implementation}
&\mathcal{L} = \frac{1}{m N} \sum_{j=1}^{m} \left ( \frac{\omega_{\mathbf{R}}}{e_{\text{POD},\mathbf{R}}} \norm{\mathbf{e}_{\mathbf{R},j}}^2 + \frac{\omega_{d}}{e_{\text{POD},d}} \norm{\mathbf{e}_{d,j}}^2 \right ) \ , \\
&\frac{\partial\mathcal{L}}{\partial\boldsymbol{\Theta}} = \frac{\partial }{\partial \boldsymbol{\Theta}} \left( \frac{2}{m N} \sum_{j=1}^{m} \left( \frac{\omega_{\mathbf{R}}}{e_{\text{POD},\mathbf{R}}} \mathbf{v}_{\mathbf{R},j} + \frac{\omega_{d}}{e_{\text{POD},d}} \mathbf{v}_{d,j} \right ) \mathbf{u}_{j} \right) \ ,
\end{aligned}
\end{equation}

\noindent where all vector quantities $\mathbf{e}_{\mathbf{R},j}$, $\mathbf{e}_{d,j}$, $\mathbf{v}_{\mathbf{R},j}$ and $\mathbf{v}_{d,j}$ are exactly as defined in Eq.~\eqref{eq:constant_vectors_definitions} and treated as constants. And the definitions of the encoder and decoder are taken from Eq.~\eqref{eq:encoder_decoder_ours}.

As we only modified the loss for the offline stage of ROM, the online stage stays untouched. Therefore, once the decoder is trained, we would proceed exactly as in Section \ref{subsec:Online_phase_of non-linear_ROMÇ_ours} with new cases to simulate.
\section{Use-Case and Evaluation Methodology}
\label{sec: usecase and evaluation methodology}

\begin{figure}
    \centering
    \includegraphics[width=\textwidth]{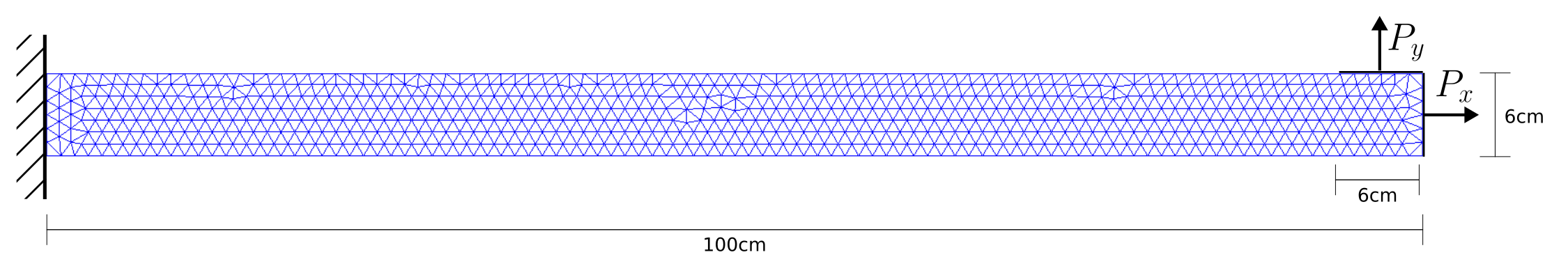}
    \caption{Mesh of the cantilever with line loads}
    \label{fig:cantilever_mesh}
\end{figure}

The evaluation of our method is done on a quasi-static structural mechanics case, specifically a non-linear case of hyperelasticity. It simulates the deformation in a 2D rubber cantilever that is fixed at its left wall and which has two different and perpendicularly-oriented line loads, $P_x$ and $P_y$, applied to its right end. The cantilever is defined by an unstructured mesh with 797 nodes, and the nodal variables to compute are the displacements in components X and Y. Therefore, our FOM system is of dimension $\mathcal{N} = 1594$. Figure \ref{fig:cantilever_mesh} shows a schematic of the described setup.

\begin{figure}
    \centering
    \includegraphics{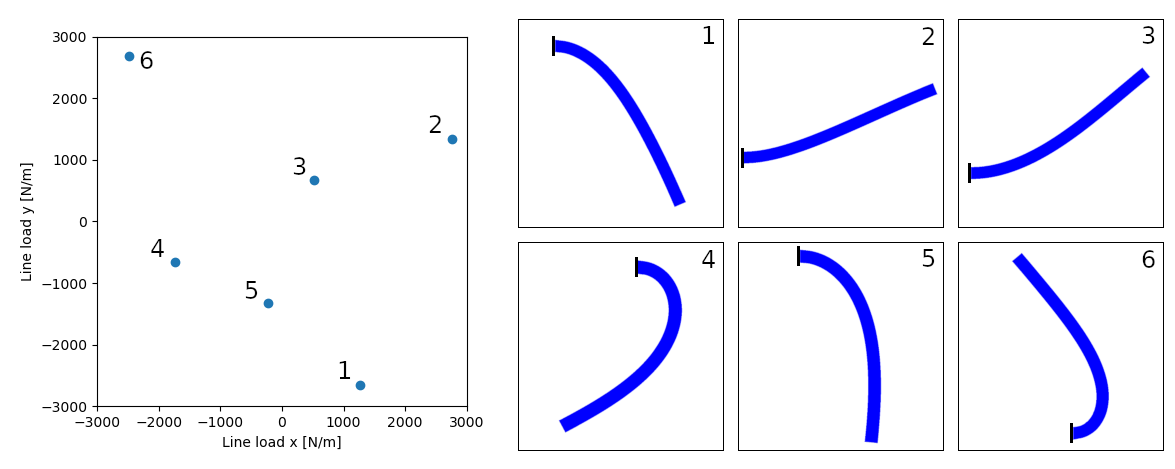}
    \caption{Examples of deformation in the cantilever for six different random combinations of line loads}
    \label{fig:samples_cantilever_examples}
\end{figure}

We define the parametric space as the range of loads that can be applied in each direction: $\mathcal{P}=[-3000,3000]\times[-3000,3000]$ N/m. And the displacements are computed in reference to the unperturbed position of each node $\bold{u}(P_x=0,P_y=0)$. Figure \ref{fig:samples_cantilever_examples} shows the deformation on the cantilever for a random set of lineload combinations within the parametric space.

In order to build the snapshot matrix for the training, we perform the ROM simulations of 5000 different cases with parameters defined by a 2D Halton pseudo-random number generator\footnote{A reduced dataset with one order of magnitude fewer FOM simulations (i.e., 500 samples) was also evaluated (see ~\ref{sec:reduced_dataset}). While this resulted in a moderate decrease in accuracy, the PROM-ANN approach remained consistently more accurate than POD. Reducing the dataset by yet another order of magnitude introduces significant challenges—particularly for training the neural network—since learning a reliable nonlinear mapping (e.g., from 10 to 40 latent coordinates) from only 50 samples is a highly underdetermined task. The choice of 5000 samples was motivated by the low cost of the benchmark and the data demands of neural network training~\cite{goodfellow2016deep}. Adaptive sampling techniques (e.g.,~\cite{liu2024application}) could be explored in future work to reduce offline cost.}. From each of these simulations, we store both the full snapshot $\mathbf{u}^*_j\in \mathbb{R}^N$ and the corresponding residual $\mathbf{R}(\mathbf{u}^*_j;\boldsymbol{\mu}) \in \mathbb{R}^N$. For the validation set we generate 1250 different samples using the same strategy, and we generate 300 more for the test set. We perform all evaluations over the test dataset; the training and the validation ones being used exclusively during the training process.

\begin{figure}
    \centering
    \includegraphics[scale=0.8]{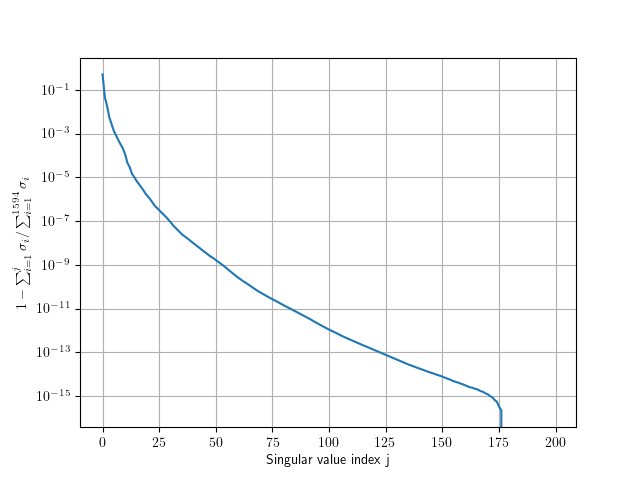}
    \caption{Singular values' energies for the first 200 modes of $\mathbf{S}_u$. }
    \label{fig:cantilever_singular_values}
\end{figure}

$\PhiInf$, $\PhiSup$, $\XiInf$ and $\XiSup$ are obtained from the SVD of the snapshots dataset for training. Figure \ref{fig:cantilever_singular_values} shows the decay of the singular values' energy for the first 200 modes of the SVD. The number of modes in each projection matrix is chosen by taking into account the accuracy obtained via traditional POD-based ROM using the same amount of modes. We establish two limit cases: one in which we select a value of $n=6$, that would lead to a $\sim0.1\%$ error on the displacements snapshot using POD, and one with $n=20$, which would achieve a relative error of $\sim10^{-6}$. Then, we take a series of $n$ values in between these to compare in our experiments. The secondary ROM basis contains a variable amount of modes so that $n+\bar{n}=60$ always.

Two different error metrics are defined:

\begin{itemize}
    \item Relative error on snapshot:
    \begin{equation}
    \label{eq:relative_snapshot_error}
    e_{u}=\exp\left(\frac{1}{M}\sum_{j=1}^M\ln\frac{\left\Vert\mathbf{u}_j-\mathbf{u}^*_j\right\Vert}{\left\Vert\mathbf{u}^*_j\right\Vert}\right)
    \end{equation}
    This is the geometric mean of the relative errors of each ROM sample $\mathbf{u}_j$ compared to the FOM one $\mathbf{u}^*_j$.
    \item Relative error on residual:
    \begin{equation}
    \label{eq:relative_residual_error}
    e_{\mathbf{R}}=\exp\left(\frac{1}{M}\sum_{j=1}^M\ln\frac{\left\Vert\mathbf{R}(\mathbf{u}_j;\boldsymbol{\mu})-\mathbf{R}(\mathbf{u}^*_j;\boldsymbol{\mu})\right\Vert}{\left\Vert\mathbf{R}(\mathbf{u}^*_j;\boldsymbol{\mu})\right\Vert}\right)
    \end{equation}
    Again, the geometric mean of the samples' relative errors, but this time comparing the residual.
\end{itemize}

We will use these metrics in the next section, in order to compare the behavior of our models both in reconstruction and online ROM.

In terms of software, we use KratosMultiphysics \cite{vicente_mataix_ferrandiz_2024_14185721} as our FEM framework, as it is open-source and lets us implement the required methods and interfaces in Python. For the neural network framework we choose TensorFlow.

All neural networks presented contain two hidden layers of 200 neurons each and with no bias. The architecture was selected based on empirical tuning through several trials using the snapshot-based loss. For the residual-based loss, the only hyperparameter modification was a reduction in the initial learning rate to prevent overwriting previously learned weights, as this stage is intended as a fine-tuning pass.
We use the Exponential Linear Unit (ELU) as activation function, following its successful application in related literature~\cite{lee2020model, BARNETT2023112420}. ELU is continuously differentiable, twice differentiable almost everywhere, and avoids vanishing gradients by allowing small negative outputs, addressing the known limitations of ReLU~\cite{lu_dying_2020}. While no explicit comparison with other activations was conducted, we believe that functions with similar smoothness and gradient-preserving properties—such as Swish~\cite{ramachandran_searching_2017}—would behave similarly in this context. We implemented ELU directly within our FEM software to allow for ANN-PROM online simulation without need for external software.
A sinusoidal learning rate scheduling strategy is applied, reducing the learning rate down to \(10^{-6}\), and the AdamW optimizer is used with TensorFlow’s default parameters. All trainings are done with batches of size \(m=16\).

We emphasize that the proposed methodology is designed and validated exclusively for forward parametric simulations.

\section{Results}
\label{sec:results}

In this section we will go over the evaluation of the methods we have proposed throughout the paper. The first subsection will study the modifications to the original PROM-ANN \cite{BARNETT2023112420} that were explained in Section \ref{sec:modification_to_PROM-ANN_architecture}. After that, we study the effect of training the architecture with the novel residual loss, as developed within Section \ref{sec:integration_of_physics-based_loss_into_PROM-ANN}. Finally we do a performance assessment.

\subsection{Modifications on the original PROM-ANN}
\label{subsec:results_modifications_on_prom-ann}

In this section, we examine the modifications introduced to the original PROM-ANN architecture during Section \ref{sec:modification_to_PROM-ANN_architecture}. The modification involved first the incorporation of scaling matrices $\XiInf$ and $\XiSup$ within the encoder-decoder pair and within the loss. Then we added a global, scalar scaling $1/e_{\text{POD},d}$ to the data-based loss. The main effects of these should be (1) achieving a similar range of values for all the neural network's inputs and (2) to make the loss and its gradients invariant to the choice of size for the latent space.

\begin{figure}
     \centering
     \begin{subfigure}[b]{0.45\textwidth}
         \centering
         \includegraphics[width=\textwidth]{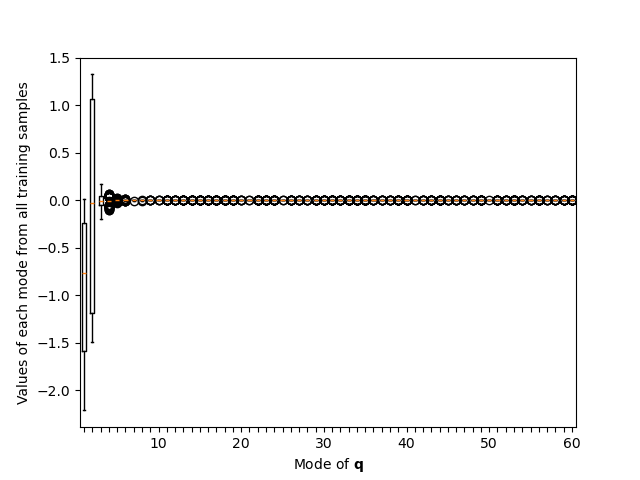}
         \caption{}
         \label{fig: boxplot q modes_a}
     \end{subfigure}
     \begin{subfigure}[b]{0.45\textwidth}
         \centering
         \includegraphics[width=\textwidth]{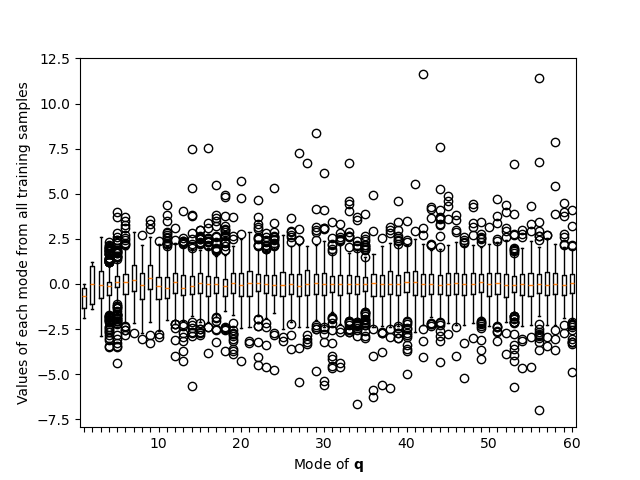}
         \caption{}
         \label{fig: boxplot q modes_b}
     \end{subfigure}
    \caption{Boxplots showing the spread of the values for each mode of $\qInf$, for all the samples in our training dataset. (a) shows the case where $\qInf = \boldsymbol{\PhiInf}^T \mathbf{u^*}$ and (b) the case where $\qInf = \XiInf^{-1}\boldsymbol{\PhiInf}^T \mathbf{u^*}$. The ranges are much more uniform in the latter case, which should be beneficial when training the neural network.}
    \label{fig: boxplot q modes}
\end{figure}

In order to confirm effect (1), we apply the encoder $E_u$ on all snapshots from our training set, with fixed latent size $n=60$. We do so with the encoder as defined in the original POD-ANN paper \cite{BARNETT2023112420} (specified in Eq.~\eqref{eq:original-prom-ann-encod-decod}) and also with ours (specified in Eq~\eqref{eq:encoder_decoder_ours}). Thus, obtaining $\mathbf{q}_{\text{orig},j}=\PhiInf^T\mathbf{u}^*_j \in \mathbb{R}^{60}$ and $\mathbf{q}_j=\XiInf^{-1}\PhiInf^T\mathbf{u}^*_j \in \mathbb{R}^{60}$, respectively, for all the snapshots. Then, in Figure \ref{fig: boxplot q modes} we plot the statistics of the obtained values for each component or mode in $\mathbf{q}_{\text{orig},j}$ and $\mathbf{q}_j$, side by side. For the original implementation (Figure \ref{fig: boxplot q modes_a}) only the first few modes have similarly large coefficients, then they become too small to differentiate. In contrast, by incorporating the singular values matrix, we obtain the result in Figure \ref{fig: boxplot q modes_b}, where all ranges are similarly large. These are, in fact, the range of values that would go into the neural network in each architecture.

\begin{figure}
     \centering
     \begin{subfigure}[b]{0.45\textwidth}
         \centering
         \includegraphics[width=\textwidth]{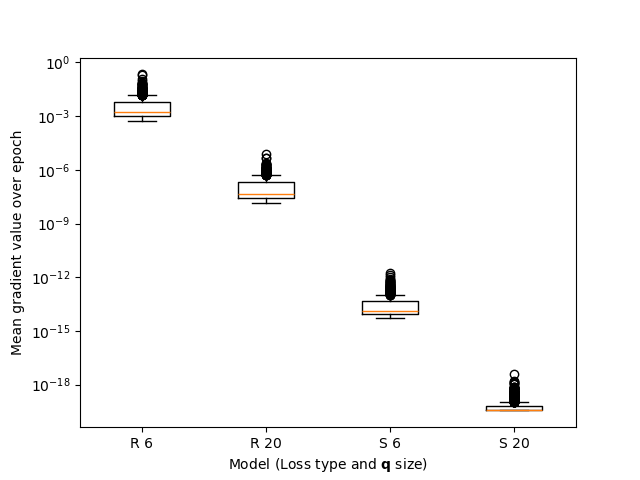}
         \caption{}
         \label{fig: boxplot gradients_a}
     \end{subfigure}
     \begin{subfigure}[b]{0.45\textwidth}
         \centering
         \includegraphics[width=\textwidth]{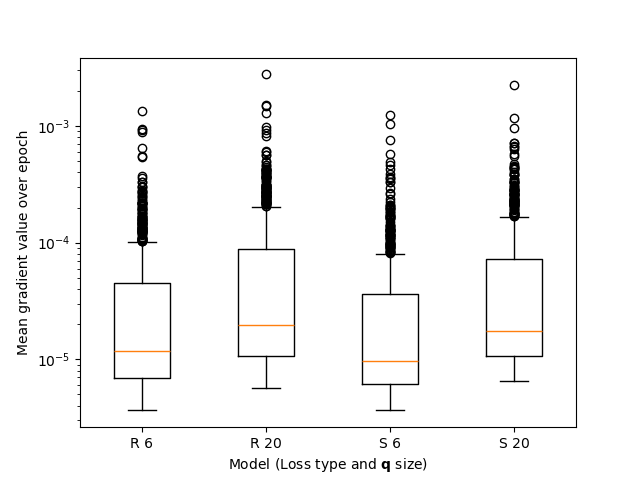}
         \caption{}
         \label{fig: boxplot gradients_b}
     \end{subfigure}
    \caption{Boxplots showing the spread of the mean value per epoch of the applied gradients during training, four different training strategies. (a) Shows the results when no rescaling of the loss is applied. (b) Shows the results when the rescaling factors $e_{\text{POD},d}$ and $e_{\text{POD},\mathcal{R}}$ are applied. The latter case is invariant to the choice of latent space size, contrary to the first one.} 
    \label{fig: boxplot gradients}
\end{figure}

Moving on to effect (2), we do a similar study but on the range of the mean of the gradients $\frac{\partial\mathcal{N}(\mathbf{q}_j;\boldsymbol{\Theta})}{\partial \boldsymbol{\Theta}}$ obtained during back-propagation in the training routine. We compute the mean gradient for each batch, and then take the mean for all batches in the training epoch. Thus, collecting a single mean gradient value per epoch. We compare the trainings with and without scaling factors $1/e_{\text{POD},d}$ and $1/e_{\text{POD},\mathbf{R}}$ (otherwise as $\mathcal{L}$ in Eq.~\eqref{eq:our_total_loss}). And for each of these two cases, we run four different training routines:
\begin{itemize}
    \itemsep0em
    \item \textit{S6}: 6 primary modes ($n=6$). Trained only on the data-based loss ($\omega_d=1$, $\omega_{\mathbf{R}}=0$).
    \item \textit{S20}: 20 primary modes ($n=20$). Trained only on the data-based loss ($\omega_d=1$, $\omega_{\mathbf{R}}=0$).
    \item \textit{R6}: 6 primary modes ($n=6$). Trained only on the residual loss ($\omega_d=0$, $\omega_{\mathbf{R}}=1$).
    \item \textit{R20}: 20 primary modes ($n=20$). Trained only on the residual loss ($\omega_d=0$, $\omega_{\mathbf{R}}=1$).
\end{itemize}
all of them for 800 epochs. We can see in Figure \ref{fig: boxplot gradients} the comparison of the statistics of these gradients for the different training modalities without the scalings (Figure \ref{fig: boxplot gradients_a}) and with the scalings (Figure \ref{fig: boxplot gradients_b}). While the former presents orders of magnitude in difference for these values depending on the size of the latent space, the latter one provides very uniform results.

The final step in the evaluation is to examine how do these modifications altogether affect the resulting decoders after training. That is, the capacity for the trained decoders to reconstruct the original snapshots, and also to serve as the approximation manifolds within ROM simulations. For this we compare three architectures and training losses:

\begin{itemize}
    \itemsep0em
    \item \textit{s-loss}: Using the encoder-decoder pair introduced by us in Eq.~\eqref{eq:encoder_decoder_ours}. Train via the data-based loss $\mathcal{L}_d$ as defined in \ref{eq:our_prom-ann_data_loss} for 800 epochs. Learning rate starting at 1e-3.
    \item \textit{q-loss}: Using the original PROM-ANN encoder-decoder pair, as in Eq.~\eqref{eq:original-prom-ann-encod-decod}. Train via the original loss $\mathcal{L}_\text{PROM-ANN}$ as defined in Eq.~\eqref{eq:original_pod-ann_loss} for 800 epochs. Learning rate starting at 1e-3.
    \item \textit{POD}: Using the fully linear POD encoder-decoder pair. Involves no training apart from the SVD computation.
\end{itemize}

\begin{figure}
    \centering
    \includegraphics[width=0.6\linewidth]{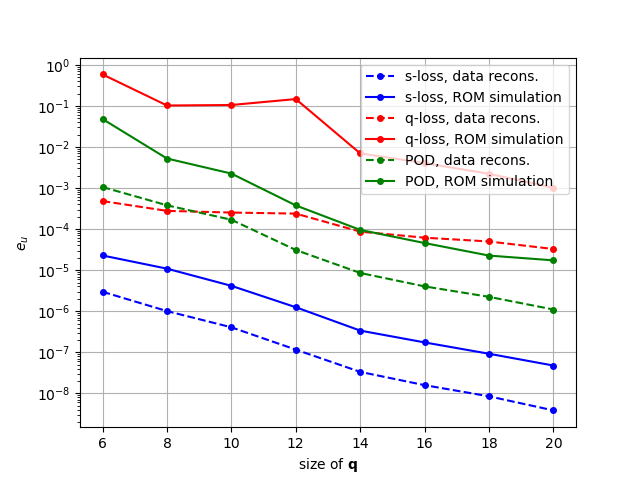}
    \caption{Comparison of the error in the snapshot, $e_u$, over three different architectures: (1) \textit{s-loss} with our modified PROM-ANN method, (2) \textit{q-loss} with the original PROM-ANN method in \cite{BARNETT2023112420}and (3) \textit{POD} with the traditional POD method. All of them for different dimensions of latent space. Dotted lines represent error on data reconstruction alone, while full lines represent error on online ROM simulation. \textit{s-loss} consistently performs several orders of magnitude better than \textit{POD}, while \textit{q-loss} is not able to keep up with \textit{POD} in most cases.}
    \label{fig:farhat_sloss_comparison}
\end{figure}

We compare the relative snapshot error ($e_u$ from Eq.~\eqref{eq:relative_snapshot_error}) applied on the reconstructed snapshots $\mathbf{u}_j=D_u(E_u(\mathbf{u}^*_j))$ for the whole test dataset. We do so for the three models specified above, and for all the $\mathbf{q}_j$ sizes specified in Section \ref{sec: usecase and evaluation methodology}. Then we do the same but taking $\mathbf{u}_j$ as the result of the online ROM simulation with each architecture. The results are illustrated in Fig. \ref{fig:farhat_sloss_comparison}.

% We first want to prove the better generalization capability of the snapshot-driven loss $\mathcal{L}_d$ in \ref{eq: loss data rom} to the one based on $q_{sup}$ proposed in \cite{BARNETT2023112420} and described in \ref{eq: loss q rom}. There were two main differences: Having the loss on error in the fully decoded snapshot $D_u(E_u(\mathbf{u*_j})$,instead of just the predicted coefficients for the superior modes ($\mathcal{N}(\mathbf{q})$).

% While being a bit more complex computationally, the architecture for the model used to train via the snapshot loss allows the inputs to the network to be better conditioned. Additionally, the loss reflects better the actual goal of the model, i.e., the snapshot reconstruction. From looking again at the spread of values in each mode at Figure \ref{fig: boxplot q modes} we see that, because of the fast decay of the range of the data for each mode, the neural network may only be effectively using the values from the first few modes.

For the data reconstruction, the error for our proposed methodology, in \textit{s-loss}, is several orders of magnitude lower than both \textit{POD} and \textit{q-loss}. While the error for \textit{q-loss} is marginally better than \textit{POD} only for the lowest sizes of $\mathbf{q}$. We attribute this to the bad conditioning of the NN input, at least for problems with fast-decaying singular values like our use-case. Without proper training, the secondary term of the PROM-ANN decoder will just add noise to the final snapshot, making it behave even worse than traditional POD.

These observations translate to a similar behavior in the ROM simulations. All the ROM simulations lose some accuracy compared to the data reconstruction, but this gap is even larger for the \textit{q-loss} case.

\subsection{Comparison of snapshot and residual losses}
\label{subsec:comparison_snapshot_and_residual_losses}

The next step is to make a comparison of the effect of training based purely on data versus training on residuals. Like the subsection before, we compare models with $\qInf$ size ranging from 6 to 20.

\begin{figure}
    \centering
    \includegraphics[width=0.6\linewidth]{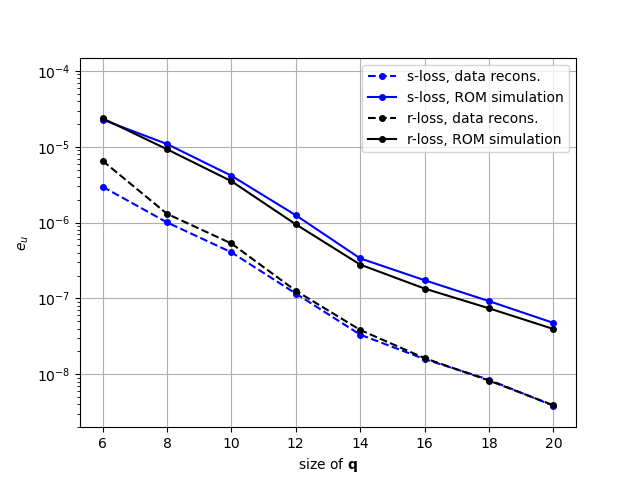}
    \caption{Comparison of the error in the snapshot, $e_u$, for models trained on loss $\mathcal{L}$ from Eq.~\eqref{eq:our_total_loss} with two different configurations: (1) \textit{s-loss} with ($\omega_d=1$, $\omega_{\mathbf{R}}=0$), from randomly initialized weights and (2) \textit{r-loss} with ($\omega_d=0$, $\omega_{\mathbf{R}}=1$) but starting from the weights of its \textit{s-loss} counterpart. Both of them for different dimensions of latent space. Dotted lines represent error on data reconstruction alone, while full lines represent error on online ROM simulation. \textit{r-loss} consistently gets slightly worse accuracy in the data reconstruction, while its ROM simulation solutions are slightly better than \textit{s-loss}.}
    \label{fig:snapshot_errors}
\end{figure}

\begin{figure}
    \centering
    \includegraphics[width=0.6\linewidth]{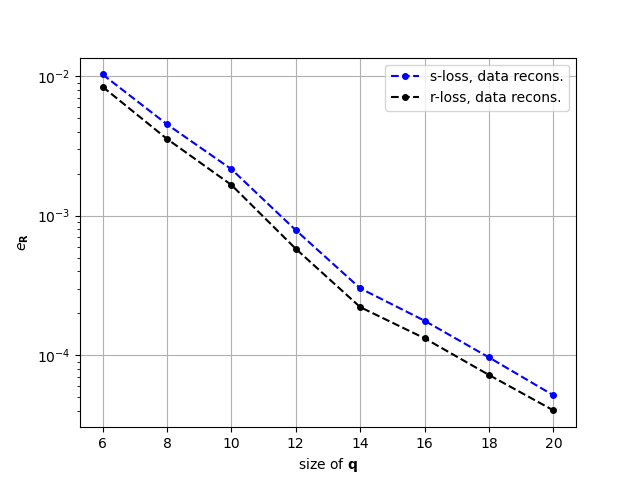}
    \caption{Comparison of the error in the residual, $e_{\mathbf{R}}$, for models \textit{s-loss} and \textit{r-loss} as described in Figure \ref{fig:snapshot_errors}. Both of them for different dimensions of latent space. The error is computed only for the case of data reconstruction. \textit{r-loss} consistently gets lower error than \textit{s-loss}.}
    \label{fig:residual_errors}
\end{figure}

Figure \ref{fig:snapshot_errors} compares errors on snapshots ($e_u$ from Eq.~\eqref{eq:relative_snapshot_error}). On the one hand, we check for the encoder-decoder data reconstruction and, on the other hand, for the results of the intrusive ROM execution. Meanwhile, Figure \ref{fig:residual_errors} compares the errors on residuals ($e_{\mathbf{R}}$ from Eq.~\eqref{eq:relative_residual_error}) only for the encoder-decoder reconstruction.

In both figures we compare two models trained differently:
\begin{itemize}
    \itemsep0em
    \item \textit{s-loss}: Train using only the snapshot loss ($\mathcal{L}$ from Eq.~\eqref{eq:our_total_loss} with $\omega_d=1$, $\omega_{\mathbf{R}}=0$) for 800 epochs.  Learning rate starting at 1e-3. (Same as in the previous sub-section)
    \item \textit{r-loss}: Start weights on the ones resulting from \textit{s-loss}. Then train for 800 epochs with only the residual loss ($\mathcal{L}$ from Eq.~\eqref{eq:our_total_loss} with $\omega_d=0$, $\omega_{\mathbf{R}}=1$). Learning rate starting at 1e-4.
\end{itemize}

The behavior represented in Figure \ref{fig:snapshot_errors} is quite interesting. Although the final result of the ROM simulation using the \textit{r-loss} model is not substantially better than its \textit{s-loss} counterpart, it consistently achieves a higher accuracy. This is in total opposition to the results for the encoder-decoder reconstruction of the snapshots, in which \textit{s-loss} gets the highest accuracy for all $\mathbf{q}$ sizes\footnote{A small set of different architecture configurations in terms of neural network layers and training batch size was tested for their accuracy in online ROM simulation (see Appendix \ref{sec:robustness_to_hyperparameter_choice}). Results did no vary significantly among different architectures, while models fine-tuned on residual-based loss consistently behaved slightly better than their data-based counterparts. This hints towards the robustness of the methodology in terms of chosen hyper-parameters, at least for the particular use-case.}. This seemingly contradictory phenomenon is clarified when looking at the reconstruction residuals in Figure \ref{fig:residual_errors} which shows the \textit{r-loss} performing consistently better. These results hint that the behavior of the intrusive ROM may be more influenced by the correctness of the residual representation for the snapshots in our ROM approximation manifold than by the accuracy of the snapshots per se.

\begin{figure}[h!]
     \centering
     \begin{subfigure}[b]{0.45\textwidth}
         \centering
         \includegraphics[width=\textwidth]{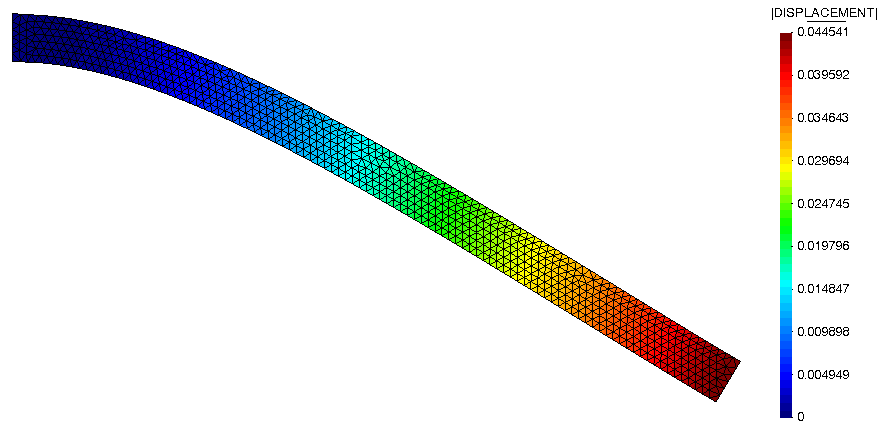}
         \caption{FOM}
     \end{subfigure}
     \begin{subfigure}[b]{0.45\textwidth}
         \centering
         \includegraphics[width=\textwidth]{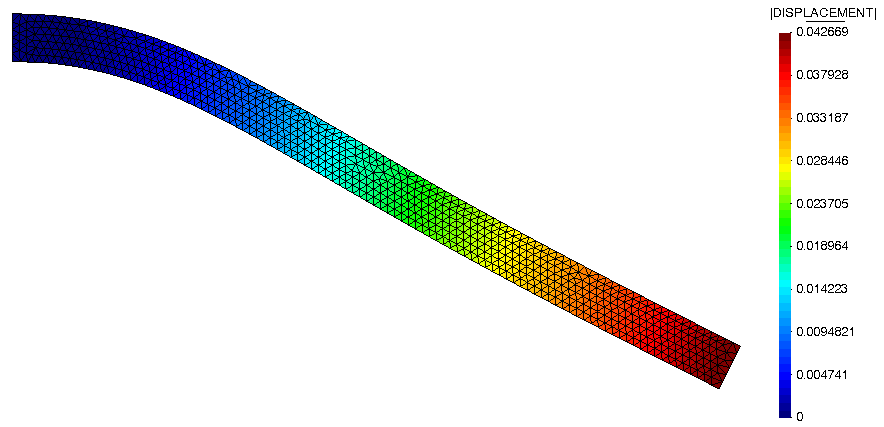}
         \caption{POD}
     \end{subfigure}
     \begin{subfigure}[b]{0.45\textwidth}
         \centering
         \includegraphics[width=\textwidth]{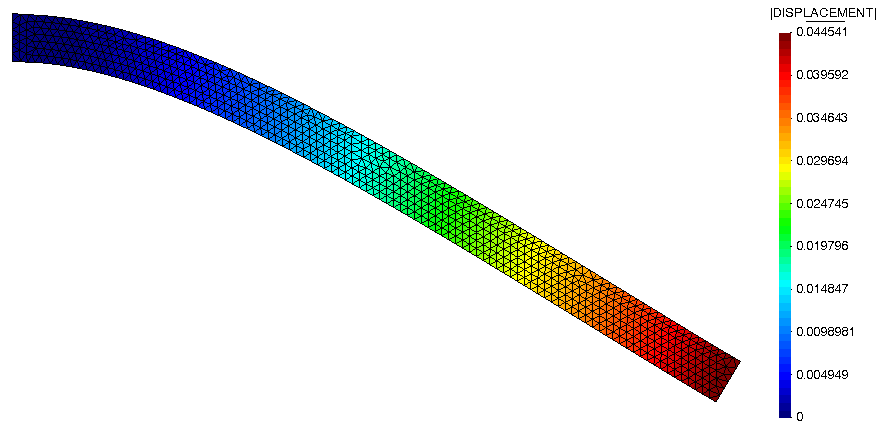}
         \caption{PROM-ANN, s-loss}
     \end{subfigure}
     \begin{subfigure}[b]{0.45\textwidth}
         \centering
         \includegraphics[width=\textwidth]{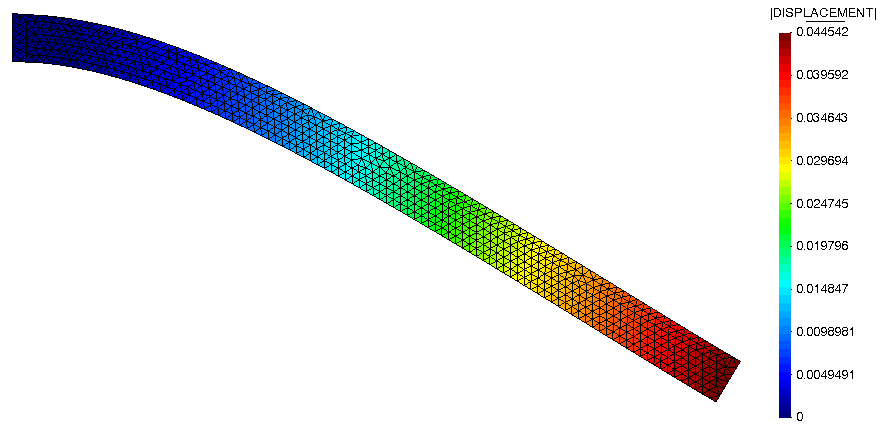}
         \caption{PROM-ANN, r-loss}
     \end{subfigure}
    \caption{Displacement representation with the results from four different aproaches, for parameter vector $\mathbf{\mu} = [2662.93,-1695.13]$. All ROM variants use $n=6$. Slight visual difference is observable between POD and FOM, while the other two models are indistinguishable from FOM.} 
    \label{fig: displacement_visualization}
\end{figure}

For visualization purposes, Figure \ref{fig: displacement_visualization} shows the displacement once applied to the cantilever for a random test case $\mathbf{\mu} = [2662.93,-1695.13]$. It does so with the results from FOM, traditional POD using $n = 6$, our PROM-ANN methodology with $n = 6$ and snapshot-based loss, and our PROM-ANN methodology with $n = 6$ and residual-based loss. As we can see, the two latter ones are virtually equal to the FOM representation, while traditional POD presents visual differences.

\subsection{Comparison of training and online ROM runtimes}
\label{subsec:comparison_of_runtimes}

For the training time, we check the mean time spent to train a single batch, measured over an entire epoch. As stted in Section \ref{sec: usecase and evaluation methodology}, all cases share the same architecture (two hidden layer of size 200) and train with batches of size 16. Computation is done on a single thread of an AMD Ryzen 7 5800x3d processor.

The description for the cases to compare is the following:
\begin{itemize}
    \itemsep0em
    \item Snapshot: Using loss $\mathcal{L}$ from Eq.~\eqref{eq:our_total_loss} with $\omega_d=1$, $\omega_{\mathbf{R}}=0$. No interaction with FEM software.
    \item Residual (optimised): $\mathcal{L}$ from Eq.~\eqref{eq:our_total_loss} with $\omega_d=0$, $\omega_{\mathbf{R}}=1$. Implementing the gradients computation using the optimizations detailed in Eq.~\eqref{eq:our_total_loss_and_gradient_implementation}.
    \item Residual (non-optimised): $\mathcal{L}$ from Eq.~\eqref{eq:our_total_loss} with $\omega_d=0$, $\omega_{\mathbf{R}}=1$. Implementing the gradients computation in a naive way by computing first $\mathbf{A}_i = (\mathbf{R}(\mathbf{u}_i)-\mathbf{R}(\mathbf{u}^*_i))^T\frac{\partial\mathbf{R}}{\partial\mathbf{u}}$ in KratosMultiphysics and then computing $\mathbf{B}_i = \frac{\partial\mathbf{u}_i}{\partial\mathbf{\Theta}}$ via the \texttt{tf.GradientTape.batch\_jacobian()} method in TensorFlow, then performing matrix multiplication in:
    \begin{equation}
    \begin{aligned}
        \frac{\partial\mathcal{L}_{\mathbf{R}}}{\partial\mathbf{\Theta}}
        =\frac{2}{e_{base,R}}\frac{1}{Nm}\sum_{i=1}^{N}\mathbf{A}_i\mathbf{B}_i
    \end{aligned}
\end{equation}
\end{itemize}

The results are printed in Table \ref{table: train_times comparison}. We see that training via the residual with the naive implementation is not feasible, as the time for a single batch is in the order of seconds. Thanks to the optimizations described in Section \ref{subsec:In-training integration of FEM software} we avoid computing entire Jacobian matrices, which reduces the time to the order of 10 milliseconds. Still, as we expected, training the neural network via the residual loss takes considerably longer than training via the snapshot loss (around 40x longer). That is why in Section \ref{subsec:comparison_snapshot_and_residual_losses} we perform a finetuning on the residual loss, instead of training from random weights.

\begin{table}[H]
    \centering
    \renewcommand{\arraystretch}{1.3} % Adjust the row height
    \begin{tabular}{|c|c|}
        \hline
        \rowcolor{gray!30}
        Loss type & Mean batch training time (s) \\ \hline
        Snapshot & 9.20e-4 \\ \hline
        Residual (optimised) & 3.87e-2 \\ \hline
        Residual (non-optimised) & 7.41 \\ \hline
    \end{tabular}
    \caption{Comparison of training time for a single batch, among different losses}
    \label{table: train_times comparison}
\end{table}

Next, we want to make a performance assessment for the online ROM execution. For that, we measure the time that the FEM software takes to solve the reduced linear system of equations. Specifically, we take the mean resolution time for all the non-linear iterations while solving 9 different cases from our test set. This is performed for two different configurations of the architecture proposed in this paper, two cases of traditional POD that achieve similar accuracies to the former pair, and the FOM simulation. Again, these evaluations have been performed using a single thread from an AMD Ryzen 7 5800x3d processor.

The results for different models can be seen in Table \ref{table: times comparison}.

\begin{table}[H]
    \centering
    \renewcommand{\arraystretch}{1.3} % Adjust the row height
    \begin{tabular}{|c|c|c|c|}
        \hline
        \rowcolor{gray!30}
        Model & $\mathbf{q}$ size & $e_u$ & System solve time (s) \\ \hline
        Modified PROM-ANN (Eq.~\eqref{eq:encoder_decoder_ours}) & 6 & 7.67e-5 & 1.42e-6 \\ \hline
        Modified PROM-ANN (Eq.~\eqref{eq:encoder_decoder_ours}) & 20 & 6.12e-8 & 5.55e-6 \\ \hline
        POD & 18 & 2.39e-5 & 4.84-6 \\ \hline
        POD & 40 & 1.32e-7 & 1.99e-5 \\ \hline
        FOM & - & - & 2.11e-3 \\ \hline
    \end{tabular}
    \caption{Comparison of computation times for various tasks within the online ROM routine, over a set of different models.}
    \label{table: times comparison}
\end{table}

% We also measure the time of performing the projection back to the full-dimensional space (and the time for updating the Right Basis in the case of PI-ANN-PROM)

% \begin{table}[H]
%     \centering
%     \renewcommand{\arraystretch}{1.3} % Adjust the row height
%     \begin{tabular}{|c|c|c|c|}
%         \hline
%         \rowcolor{gray!30}
%         Model & $e_u$ & System solve time (s) & Back-projection time (s) \\ \hline
%         DPI-ANN-APROM (6.60) & 7.67e-5 & 1.42e-6 & 1.70-4 \\ \hline
%         DPI-ANN-APROM (20.60) & 6.12e-8 & 5.55e-6 & 2.78e-4 \\ \hline
%         POD (18) & 2.39e-5 & 4.84-6 & 8.97e-5 \\ \hline
%         POD (40) & 1.32e-7 & 1.99e-5 & 9.76e-5 \\ \hline
%         FOM & - & 2.11e-3 &  - \\ \hline
%     \end{tabular}
%     \caption{Comparison of computation times for various tasks within the online ROM routine, over a set of different models.}
%     \label{table: times comparison}
% \end{table}

As expected, the main computational advantage of our method (and of the original PROM-ANN \cite{BARNETT2023112420}, as both architectures are equivalent in terms of computational complexity) comes from the achieving smaller linear systems to solve for the same precision. It is noteworthy to point out that the biggest gains in computation time for the online process come with a further step in the ROM workflow called Hyper-Reduction, in which the amount of elements to take into account for the generation of the system at each step is greatly reduced. The number of reduced elements to use is directly related to the latent space dimension. Therefore, the reduction in modes that we achieve is also beneficial in this regard. This study focuses on the offline part of the ROM process, so there is no further development on the online procedure compared to the original PROM-ANN method. The implementation of Hyper-Reduction is described for this type of architecture in their paper \cite{BARNETT2023112420}, so interested readers are encouraged to check their performance assessments, which should be applicable to our case aswell.
% \section{Discussion}
% \begin{itemize}
%     \item Interpretation of the results, their implications, and potential areas of improvement.
%     \item Discussion on the broader impact of the PI-ANN-APROM approach in the field of computational mechanics.
% \end{itemize}

% Therefore, while the results for the residual loss are not overwhelming (taking account the increase in training time it supposes), these insights set a precedent for more efforts in investigating the role of the intrinsic residual in the solutions manifold and how to build it taking this aspect into account.

\section{Discussion and future work}
\label{sec:discussion_and_future_work}

Once we have assessed the performance of our proposed architecture and losses, we will further discuss the implications of these results and the impact of our contributions.

We comment first on the discrete FEM residual-based loss that we developed throughout Section \ref{sec:discrete_pinn}, and its proposed implementation strategy. These developments pose a step-up from recent initiatives to design discrete PINN-like losses \cite{Meethal2023, Thang2023, yamazaki_finite_2025, sunil_fe-pinns_2024} that limit themselves to linear problems of different natures. By taking advantage of open-source FEM software like KratosMultiphysics \cite{vicente_mataix_ferrandiz_2024_14185721}, we can access all the FEM methodology needed to obtain residuals and Jacobians for non-linear cases in a wide range of state-of-the-art FEM formulations. The cost for this is a loss in time efficiency during training, compared to the classic data-based approach, because of the required dynamic interaction between the FEM software and the neural network framework during training. And also because of the manual computation of the gradient loss. It is in this sense that our implementation proposal makes a huge difference, taking the training time from being prohibitive to being just an order of magnitude higher than the data-based one. These comparisons are shown in Section \ref{subsec:comparison_of_runtimes}. We believe that the main computational bottleneck after our proposed methodology is the fact that the FEM software currently runs the computations in series for each sample in the batch. Some future work on parallelizing these procedures could potentially unlock training times much closer to the data-based ones. In addition, the current formulation introduces a fundamental architectural shift compared to the original PROM-ANN framework~\cite{BARNETT2023112420}, which operated entirely in the reduced-order space, learning a map from \( q \in \mathbb{R}^n \) to \( \bar{q} \in \mathbb{R}^{\bar{n}} \). In contrast, our physics-aware variant requires training in the high-dimensional physical space of the full-order model, since the FEM residual and its Jacobian must be evaluated in that space. This change increases training cost considerably, both due to the FEM evaluations and the need to retrieve full Jacobians for backpropagation. While this enables the integration of high-fidelity physics, it does not scale well for large-scale problems. Future work could address this by coupling the current approach with scalable network architectures—e.g., convolutional or graph-based neural networks—and
by projecting the residual into an intermediate reduced-order space before backpropagation. Related ideas have emerged in recent literature, notably in the form of semi-intrusive training strategies such as the one proposed by Halder et al.~\cite{halder2023physics}, where residuals are used only in projected form during training, avoiding full-order evaluations. However, these approaches preserve non-intrusiveness. We believe such hybrid strategies are promising and plan to explore them in future developments. Another key characteristic of the proposed loss is being parameter-agnostic. This makes it more versatile in various ways: in terms of efficiency the FEM software does not need to re-configure the simulation for each specific sample, and in terms of use-case it can be applied to cases in which not only the minimization of the residual itself is important, but also its behavior while being non-zero. This latter aspect is key in using this loss for our particular setting of intrusive ROM.
One unexplored advantage of this loss formulation would be the possibility to perform partial physics-based learning, where only specific components of the total residual are used for the training. For example one could train only on the steady-state component of the residual of a dynamic case in order to avoid the inconveniences from the dependency on previous time-steps. This is not addressed in this paper, but left as possible future contributions. There are further options that the residual loss could open up and that we haven't fully explored, e.g. the possibility of training on the residual with noisy data in order to achieve data augmentation.
Next, we comment on the modification to the PROM-ANN architecture itself and the data-based loss $\mathcal{L}_d$ with the scaling matrices $\boldsymbol{\Xi}$, $\boldsymbol{\bar{\Xi}}$ and the global scaling $1/e_{\text{POD},d}$. The scaling matrices are an inexpensive way to normalize the input ranges for the neural network with the direct results from the SVD, so we avoid extra statistical studies of the dataset. The global scaling is a single scalar computed inexpensively via POD, only once for the whole training. The effect of these modifications is apparent in the results in Section \ref{subsec:results_modifications_on_prom-ann}, where both the reconstruction and ROM results from our architecture are several orders of magnitude better than the simpler, original approach described in \cite{BARNETT2023112420}. Now, there is a very plausible explanation for the lack of performance of the original PROM-ANN, mostly when comparing with the good results that they obtain in their paper. The use case that they use for evaluation is a 2D inviscid Burgers problem, which has a much flatter decay in the SVD's singular values compared to ours. Thus, the range of their inputs to the neural network should naturally be more uniform. Another possibility is that they apply some normalization routine prior to the neural network without mentioning it explicitly. In any case, we can say that our modification makes the architecture generalizable to any kind of problem in terms of their singular value decay. Additionally, the global scaling $1/e_{\text{POD},d}$ is key in order to stabilize the scale of the loss and the backpropagation gradients, making the neural network optimizer perform equally whatever the choice of latent size. It also has an interpretational purpose, which is to make the loss a direct indicator of how much better the results are relative to the simple POD version. Something to explore in the future would be better methodologies to choose the modes included in the primary and secondary ROBs. Right now this is done in a greedy way, selecting as many modes are needed to achieve a certain accuracy in POD, but it is not clear how different modes are related to one another (especially since they are uncorrelated in the linear sense). The implementation of our version of ANN-PROM (only with the data-based loss) is readily available within the ROM Application of KratosMultiphysics \cite{vicente_mataix_ferrandiz_2024_14185721}\footnote{The framework to train via residual is not yet implemented in the master branch of the KratosMultiphysics software at the time of publication. Until it is fully implemented, interested users can check out the branch at \url{https://github.com/KratosMultiphysics/Kratos/tree/RomApp_RomManager_ResidualTrainingStructural}, which handles the residual usage for the StructuralMechanicsApplication. Once using that branch, the user may run the example in \url{https://github.com/KratosMultiphysics/Examples/tree/master/rom_application/RomManager_cantilever_NN_residual}, which implements this cantilever use-case (with reduced number of samples by default).}

Finally, we enter the discussion about the effect of training our modified ANN-PROM architecture on the residual loss. Other than the increased duration of the training routine, it was also difficult for us to prevent the model from falling in non-optimal local minima during training with the residual loss. That is why the chosen approach was to train first with the data-based approach and only then finetune with a purely physics-based loss. The intuition for proposing this physics-based training is that the intrusive ROM accuracy is not only given by the snapshot-reconstruction capabilities of the approximation manifold (which would work more like a lower limit in terms of error), but should also be dependant on how well the residuals are represented within these solutions in the manifold. Essentially because the residual is the quantity being optimized during the intrusive ROM simulation. The general discrepancy between ROM and reconstruction is clearly demonstrated within our use-case in Section \ref{subsec:results_modifications_on_prom-ann}, with the error in ROM simulation being approximately one order of magnitude higher than the reconstruction one for both our proposed architecture with the data-based loss and traditional POD. Further on we look at the results in Section \ref{subsec:comparison_snapshot_and_residual_losses} to specifically understand the effect of the residual-based training. The observations are encouraging, even if not spectacular. We say this in the sense that performing ROM with the model trained on the residual provided slightly but consistently better results than the one trained with the snapshot. It can also be interpreted as achieving a slightly lower discrepancy between reconstruction and ROM simulation, which is what we were aiming for. But the most important insight is that this phenomenon coincides with a consistently lower error in the representation of the residual by the models trained on physics. We are aware that, as it is right now, the significantly higher training time with the residual loss renders the proposed method not attractive, taking into account the marginal increase in ROM accuracy, but we are optimistic that future research can enable more meaningful improvements.

We make the observation, in retrospective, that choosing the ANN-PROM architecture for implementing the residual loss limited the achievable accuracy. That is because, even if the neural network allows us to introduce the loss of our choosing, we still depend entirely on the modes that we gathered via SVD on our snapshots dataset, without any regard for the residual accuracy. The fact that even with this caveat we were able to achieve slightly better results makes us very optimistic for the future where we could explore new architectures (or variants of this one) that put the residual in the focus point from the beginning, or that allow more freedom to correct the residual on top of the modes for the snapshot. This restriction of ANN-PROM towards the flexibility of the residual also hinders the potential ability of the residual training to enhance the model's behavior outside the training parameter space, which is a topic worthy of studying in detail in further work (see ~\ref{sec:extrapolation_ability}).

Although the present work focuses solely on augmenting the PROM-ANN framework with a discrete residual loss to enhance physics-awareness, it is useful to briefly comment on its positioning within the broader landscape of nonlinear ROMs. In this paper, comparisons were limited to traditional linear subspace PROMs, as our contribution centers specifically on the integration of the high-fidelity residual in the PROM-ANN training. Broader comparisons with alternative nonlinear ROM techniques—such as quadratic ROMs or kriging-based interpolation—relate more directly to the PROM-ANN methodology itself, independently of the residual loss term. In this context, recent studies such as~\cite{chmiel2024assessment} have benchmarked PROM-ANN against several nonlinear ROM strategies. Furthermore, the residual-based loss proposed in this work is not restricted to PROM-ANN and can be readily integrated into other neural-network-based ROM architectures, such as convolutional autoencoder ROMs~\cite{lee2020model}. It is also fair to note that, within the broader family of nonlinear PROMs, our formulation remains fully compatible with local or piecewise PROM-ANN variants—whether using a single ANN that adapts to the active local basis, or multiple models trained for different regions in parameter space. These broader methodological directions are considered valuable future extensions.

All in all, the three main contributions of the paper work together to complement each other and obtain a physics-informed intrusive ROM framework, but also hold value by themselves: the residual loss could be applied for other purposes like non-intrusive ROM, the modifications to the ANN-PROM architecture make it more versatile in terms of the types of problems it can handle, and the study of the effect of the residual in intrusive ROM provides a seed for a new path of research in this discipline for us and for the rest of the community in numerical methods.

\section{Conclusions}
\label{sec:conclusions}

In this paper, we extend the PROM-ANN architecture proposed in \cite{BARNETT2023112420} by incorporating a training approach based on the finite element method (FEM) residual, rather than relying solely on snapshot data. This establishes a connection between non-linear reduced-order models (ROMs) and physics-informed neural networks (PINNs). While traditional PINNs use analytical partial differential equations (PDEs) to train continuous, non-intrusive models, our approach leverages discrete FEM residuals as the loss function for backpropagation, guiding the learning of the ROM approximation manifold. This development allows us to investigate the impact of improving the residual of the snapshots on the overall performance of projection based ROMs.

The path to achieve this final goal enables us to present three independently significant contributions: (1) a loss based on the FEM residual and which is parameter-agnostic and, most importantly, applicable to non-linear problems; (2) A modification on the original PROM-ANN architecture in \cite{BARNETT2023112420} that makes it applicable to cases with fast-decaying singular values, and (3) a study on the effect of the residual-based training on ROM simulation. We demonstrate our approach in the context of static structural mechanics with non-linear hyperelasticity, specifically in the deformation of a rubber cantilever subjected to two orthogonal variable loads. 

In terms of the residual loss, the fact that it is based on existing FEM software makes it applicable to a high range of problems. The proposed implementation strategy makes the interaction between the neural network and the FEM software viable in reasonable training time, around 40ms per batch.  Finally, the enhancement of the resulting residuals via this loss is demonstrated by applying it on our proposed PROM-ANN-based architecture and performing data reconstruction. This results in a consistently lower residual representation compared to the cases trained on snapshot data alone.

About the modifications to the original ANN-PROM architecture, we observe how our method lowers the snapshot error by several orders of magnitude compared to POD, in both the data reconstruction and the ROM simulation results. This improvement is consistent for latent space sizes ranging from $n=6$ to $n=20$. In contrast, the original PROM-ANN formulation struggles to train the neural network, resulting in errors higher than POD in most cases. This enhancement from our methodology comes from a proper scaling strategy on both the architecture itself and the loss to be used.

Finally, for the effect of applying the FEM residual loss on the approximation manifold for projection-based ROM, our results show a modest but consistent reduction of the gap between the accuracy for snapshot reconstruction and the accuracy for projection-based ROM. While not being useful in practice as of now, this observation makes us optimistic that better results for projection-based ROM could be unlocked in the future by taking care of the residual representations within the approximation manifold. In the future, alternative non-linear ROM architectures that enable more control over the resulting solutions' residuals can greatly improve the results presented in this work.

As a corollary to the discussion and to provide a roadmap for future work, we summarize below the main current limitations of the proposed approach and potential directions to address them:

\begin{table}[h]
\centering
\renewcommand{\arraystretch}{1.2}
\begin{tabular}{|p{0.44\textwidth}|p{0.44\textwidth}|}
\hline
\textbf{Current Limitation} & \textbf{Potential Direction} \\
\hline
High training cost due to serial FEM residual evaluations. & Parallelize FEM evaluations across mini-batches to improve training throughput. \\
\hline
Residual evaluations are performed in full-order space, limiting scalability. & Project residuals into intermediate reduced-order spaces to reduce computational cost.\\
\hline
Residual-based training yields only modest gains in ROM accuracy. Attributed to PROM-ANN being limited by fixed SVD-based modes and not being able to adapt to residual structure. & Design residual-aware or adaptive mode selection strategies. Architectures like convolutional or graph neural networks may be good candidates for their flexibility, but may be limited in terms of scalability.\\
\hline
Extrapolation remains limited due to lack of residual-focused manifold generalization. & Explore local or multi-network manifold strategies; study residual-informed extrapolation systematically. \\
\hline
Method limited to forward problems & Explore inverse problems via adjoint-based gradients, or parameter-to-output mappings \\
\hline
\end{tabular}
\caption{Summary of current limitations and future research directions for the proposed discrete physics-informed residual loss for projection-based ROMs.}
\label{tab:limitations_summary}
\end{table}

\section{Nomenclature}

\begin{tabular}{ll}
    HFM & high-fidelity model \\
    ROM & reduced order model \\
    POD & proper orthogonal decomposition \\
    SVD & singular value decomposition \\
    PINN  & Physics Informed Neural Network \\ 
    PDE & partial differential equation \\
    FEM & finite elements method \\
    $\mathcal{N}: (\boldsymbol{a} ; \boldsymbol{\Theta} ) \mapsto \boldsymbol{b}$ &  a neural Network parametrized by $\boldsymbol{\Theta}$ \\
    $\boldsymbol{u}$ & nodal solution of FEM problem \\
    $\boldsymbol{q}$ & FEM solution representation in a given reduced space \\
    $D : \boldsymbol{q} \mapsto \boldsymbol{u}$ & a decoder function \\
    $E : \boldsymbol{u} \mapsto \boldsymbol{q}$  & an encoder function \\
    $\boldsymbol{\mu}$ & parameters vector for a FEM simulation \\
    $\mathbf{R}(\boldsymbol{u};\boldsymbol{\mu})$ & FEM residual, given nodal solution $\boldsymbol{u}$ and simulation parameters $\boldsymbol{\mu}$\\
    $\mathcal{L}(\boldsymbol{\Theta})$ & loss function used to optimize parameters set $\boldsymbol{\Theta}$ \\
    $N \in \mathbb{N} $ & number of degrees of freedom in the FOM\\
    $n \in \mathbb{N}$  & number of degrees of freedom in ROM. Number of POD modes. Latent space dimensions\\  
    $M \in \mathbb{N}$  & number of samples in the training dataset\\
    $m \in \mathbb{N}$  & number of samples in a given batch when training a neural network\\
\end{tabular}

\section*{Acknowledgements}
N. Sibuet acknowledges the Secretariat of
Universities and Research of the Department of Research and Universities of the
Generalitat of Catalonia, as well as the European Social Plus Fund for their financial support through the predoctoral scholarship AGAUR-FI (2024 FI-3 00065) Joan Oró.

S. Ares de Parga and J.R. Bravo acknowledge the Departament de Recerca i Universitats de la Generalitat de Catalunya for the financial support through the FI-SDUR 2020 and FI-SDUR 2021 scholarships.

S. Ares de Parga also acknowledges support from the Fulbright Commission Spain through a Fulbright Predoctoral Research Fellowship (2024–2025).

\bibliographystyle{unsrt}
\bibliography{sample.bib}

\appendix
\section{Reduced dataset}
\label{sec:reduced_dataset}

\begin{figure}[h!]
\centering
\includegraphics[width=0.6\textwidth]{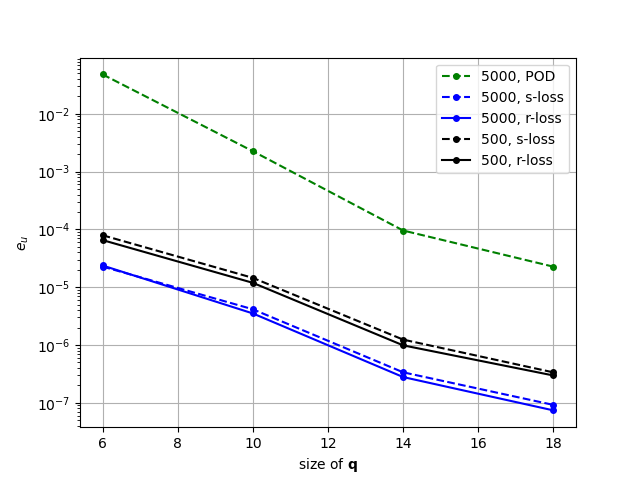}
\caption{Comparison of the relative error \( e_u \) versus the reduced coordinate \( \textbf{q} \) size for POD and PROM-ANN variants trained with 500 and 5000 FOM simulations. PROM-ANN consistently outperforms POD, and residual-based training leads to improved accuracy in both sample regimes.}
\label{fig:promann_pod_qsize}
\end{figure}

This appendix shows the details for the performance of our methodology while using a reduced training dataset of 500 FOM samples. The compared models are those from Section \ref{subsec:comparison_snapshot_and_residual_losses} with 6, 10, 14 and 18 primary modes, which were trained with 5000 samples, and their equivalent counterparts trained in the reduced dataset. Figure \ref{fig:promann_pod_qsize} shows how the reduced dataset leads to a moderate degradation in accuracy, but the models are still significantly more accurate than the POD baseline, both with and without the proposed residual-based training. This confirms that, for many applications, 500 samples may already offer a satisfactory trade-off between offline cost and prediction quality.

\section{Extrapolation ability}
\label{sec:extrapolation_ability}

\begin{figure}[h!]
\centering
\includegraphics[width=0.6\textwidth]{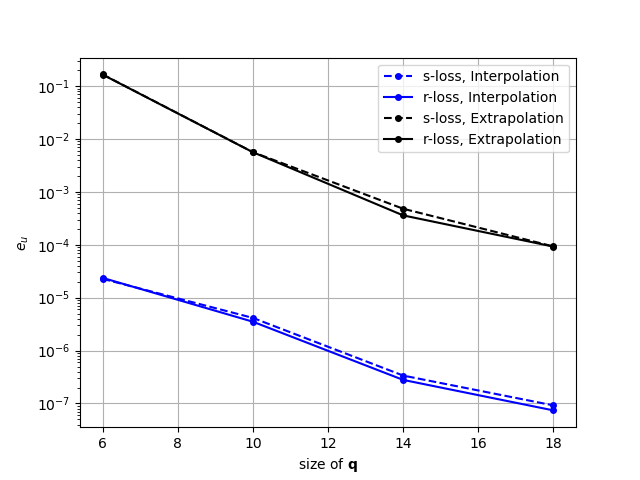}
\caption{Comparison of the relative error \( e_u \) versus the reduced coordinate \( \textbf{q} \) size for PROM-ANN variants evaluated either in (1) the original test set (Interpolation): all samples contained within the parametric space, or (2) an Extrapolation set: 50 samples outside the parametric space, within 500N/m off the limits. While the accuracy is severely decreased compared to interpolation, the extrapolation performance is still reasonable for the cases with higher \( \textbf{q} \) size.}
\label{fig:promann_extrapolation_qsize}
\end{figure}

This appendix shows the results of testing models resulting from our methodology outside of the training parameter space. Specifically, we generate a test set of 50 FOM samples outside the parametric space, within 500N/m off the limits. Then we test the same models as in Section \ref{subsec:comparison_snapshot_and_residual_losses} (for 6, 10, 14 and 18 primary modes) on this new test set and compare them to the results on the original test dataset. Figure \ref{fig:promann_extrapolation_qsize} shows how, while significantly inferior to interpolation capabilities, when using a higher number of primary modes the
capacity for extrapolation becomes reasonable, achieving errors lower than 1e-3.

\section{Robustness to hyper-parameter choice}
\label{sec:robustness_to_hyperparameter_choice}

This last appendix shows the results of training a few extra ANN-PROM models with different neural network layer configurations or batch sizes. All of these new models use \(n = 14\) and \(\bar{n} = 46\), and both the snapshot-based and residual-based models are represented. They have all been evaluated for online snapshot accuracy $e_u$.

The gathered accuracies, listed in Table \ref{table: hyper-parameters comparison}, do not vary significantly from one architecture to another. On top of that, models finetuned on the residual-based loss still consistently have a slight advantage over their data-based trained peers. Therefore, the results indicate that the proposed methodology is indeed applicable to various configurations, at least in the use-case at hand.

\begin{table}[H]
    \centering
    \renewcommand{\arraystretch}{1.3} % Adjust the row height
    \begin{tabular}{|c|c|c|c|}
        \hline
        \rowcolor{gray!30}
        Layers size & Batch size & \(e_u\) (s-loss) & \(e_u\) (r-loss) \\ \cline{1-4}
200         & 16         & 7.43e-07      & 6.49e-07  \\ \hline
200, 200    & 8          & 4.31e-07      & 3.24e-07  \\ \hline
200, 200    & 16         & 3.37e-07	     & 2.79e-07  \\ \hline
400, 400    & 16         & 4.09e-07	     & 3.26e-07  \\ \hline
    \end{tabular}
    \caption{Comparative study of different neural network architecture configurations (all with \(n = 14\) and \(\bar{n} = 46\)). The relative error on snapshot \(e_u\) from online ANN-PROM simulation is shown for models trained on data only (s-loss) and models finetuned on residual loss (r-loss). Performance does not not change significantly among different architectures, but r-loss models still consistently have a small advantage over s-loss.}
    \label{table: hyper-parameters comparison}
\end{table}

\end{document}